\documentclass[11pt]{article}

\usepackage[final]{acl}
\usepackage{xcolor}
\usepackage{colortbl}  
\usepackage{amsmath}
\usepackage{wasysym}

\usepackage{times}
\usepackage{latexsym}
\usepackage{booktabs}

\usepackage{times}
\usepackage{latexsym}

\usepackage[T1]{fontenc}

\usepackage[utf8]{inputenc}

\usepackage{microtype}
\usepackage{pifont}
\usepackage{inconsolata}

\usepackage{makecell}
\usepackage{xcolor}
\usepackage{wasysym}
\usepackage{times}
\usepackage{latexsym}
\usepackage{booktabs}
\usepackage[T1]{fontenc}

\usepackage[utf8]{inputenc}

\usepackage{microtype}
\usepackage{multicol}
\usepackage{multirow}
\usepackage{inconsolata}
\usepackage{graphicx}

\usepackage{tcolorbox}

\newtcolorbox{myprompt}[2][]
{
    colframe=black,      
    colback=gray!20,     
    boxrule=1pt,         
    arc=4pt,             
    left=10pt,           
    right=10pt,
    top=10pt,            
    bottom=10pt, 
    title=#2
}
\title{HyperGVL: Benchmarking and Improving Large Vision-Language Models in Hypergraph Understanding and Reasoning}

\author{
\textbf{Yanbin Wei}$^{1,2}$\thanks{Equal contribution.} \quad
\textbf{Chun Kang}$^{4}$\footnotemark[1] \quad
\textbf{Siwei Li}$^{4}$ \quad
\textbf{Haoxuan Che}$^{3}$ \quad
\textbf{Yang Chen}$^{1}$ \quad
\textbf{Hua Liu}$^{1}$ \quad
\textbf{Jian Liu}$^{4}$ \quad \\
\textbf{Zhuang Liu}$^{4}$ \quad 
\textbf{Can Ouyang}$^{4}$ \quad
\textbf{Fei Xing}$^{1}$ \quad
\textbf{Lei Sha}$^{4}$\thanks{Corresponding author.} \quad 
\textbf{Rui Liu}$^{3}$\footnotemark[2] \quad
\textbf{Yu Zhang}$^{1}$\footnotemark[2] \quad
\textbf{James Kwok}$^{2}$ \\
$^1$Southern University of Science and Technology\quad\\
$^2$Hong Kong University of Science and Technology\\
$^3$Huawei Research \quad
$^4$Beihang University\\
}

\begin{document}
\maketitle

\begin{abstract}
Large Vision-Language Models (LVLMs) consistently require new arenas to guide their expanding boundaries, yet their capabilities with hypergraphs remain unexplored. In the real world, hypergraphs have significant practical applications in areas such as life sciences and social communities. Recent advancements in LVLMs have shown promise in understanding complex topologies, yet there remains a lack of a benchmark to delineate the capabilities of LVLMs with hypergraphs, leaving the boundaries of their abilities unclear. To fill this gap, in this paper, we introduce \texttt{HyperGVL}, the first benchmark to evaluate the proficiency of LVLMs in hypergraph understanding and reasoning. \texttt{HyperGVL} provides a comprehensive assessment of 12 advanced LVLMs across 84,000 vision-language question-answering (QA) samples spanning 12 tasks, ranging from basic component counting to complex NP-hard problem reasoning. The involved hypergraphs contain multiscale synthetic structures and real-world citation and protein networks. Moreover, we examine the effects of 12 textual and visual hypergraph representations and introduce a generalizable router \texttt{WiseHyGR} that improves LVLMs in hypergraph via learning adaptive representations. We believe that this work is a step forward in connecting hypergraphs with LVLMs.
\end{abstract}

\section{Introduction}
Graphs serve as a fundamental data structure for modeling the relationships between abstract concepts or tangible objects in the real world. 
Among the sub-categories of graphs, hypergraphs are significant because their hyperedges can effectively model high-order correlations among three or more entities. The applications of hypergraphs are prevalent in the real world.  
For example, in social networks,
hypergraphs can naturally represent community interactions, where a hyperedge can connect an arbitrary number of vertices, reflecting the complex, high-order relationships within communities \cite{contisciani2022inference}.
Similarly, in life science, 
hypergraphs are adept at modeling interactions such as catalytic triplets in protein structures
\cite{ravetz2019photoredox}, 
where ordinary graphs that only focus on pair-wise relations fall short. Recent advances 
also shows the promising  capability of
hypergraph in  modeling complex relationships among information in retrieved-augmented generation \cite{feng2025hyperrag} more accurately.

\begin{figure}[t]
    \centering
    \includegraphics[width=\linewidth]{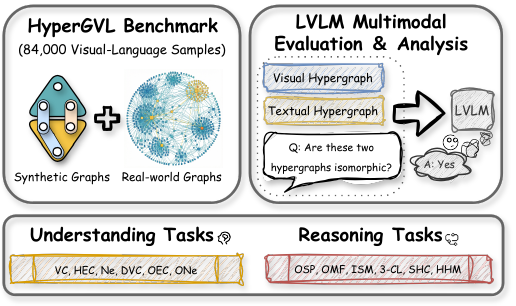}
    \vspace{-0.7cm}
    \caption{Overview of the \texttt{HyperGVL} benchmark.}
    \label{fig:overview}
    \vspace{-0.5cm}
\end{figure}

\begin{table*}[!htbp]
\centering
\resizebox{\linewidth}{!}{
\begin{tabular}{lcccccc}
\toprule
\textbf{Benchmark}   & \textbf{Evaluation Capability}   & \textbf{Graph Source} & \textbf{Structure Perception} & 
\textbf{High-Order} & 
\textbf{\#Tasks} &
\textbf{\#Samples} \\
\midrule
\textbf{GVLQA}  & Reasoning    & Synthetic   & Visual \& Textual & \textcolor{red}{\ding{55}} & 7   & 157,896           \\
\textbf{VisionGraph} & Understanding \& Reasoning  & Synthetic & Visual &\textcolor{red}{\ding{55}}   & 10 & 3,000           \\
\textbf{VGCure}  &  Understanding \& Reasoning    & Synthetic \& Real-world  &   Visual & \textcolor{red}{\ding{55}}  & 22 & 223,646\\

\textbf{LLM4Hypergraph} & Mainly Understanding & Synthetic \& Real-world  &   Textual & \textcolor{green}{\ding{51}} & 15 & 21,500 \\
\midrule
\textbf{\textit{HyperGVL (Ours)}} &   Understanding \& Reasoning    & Synthetic \& Real-world  & Visual \& Textual & \textcolor{green}{\ding{51}} & 12 & 84,000 \\

\bottomrule
\end{tabular}}
 \vspace{-0.2cm}
\caption{Comparisons between \texttt{HyperGVL} and related graph analysis benchmarks for LVLMs/LLMs. \textit{\#Tasks}: number of response types; \textit{\#Samples}: total number of test samples; \textcolor{green}{\ding{51}}/\textcolor{red}{\ding{55}}: support/not support high-order relationships.}
\label{tab:all_bench}
\vspace{-10pt}
\end{table*}
On the other hand, large vision-language models (LVLMs) exhibit outstanding performance
across a wide range of downstream tasks with human-like understanding
and reasoning abilities \cite{li2025benchmark}. This triggers a
growing interest in employing LVLMs for graph
learning problems, as the vision modality offers a
natural way for comprehending structural information and facilitating graph-related
reasoning, where  \texttt{GVLQA} \cite{wei2024gita}, \texttt{VisionGraph} \cite{li2024visiongraph}, and \texttt{VGCure} \cite{vgcure} are among the first batch of such methods. However, they limit their scopes to ordinary graphs, and do not explore the potential of LVLMs on high-order relationships in hypergraphs.

To address this gap, we introduce \texttt{HyperGVL} (Fig. \ref{fig:overview}), the first comprehensive benchmark dataset designed to evaluate the capabilities of LVLMs on hypergraphs. \texttt{HyperGVL} consists of 84,000 vision-language question-answering (QA) pairs, covering both multiscale synthetic hypergraphs and real-world hypergraphs from citation and protein networks.  The evaluation spans 12 tasks of varying difficulty levels, from fundamental hypergraph component understanding to the challenging NP-hard problems reasoning. Additionally, \texttt{HyperGVL} integrates seven textual and five visual representations of hypergraphs, offering insights into task preferences and model capability boundaries across these diverse representations. 

Based on the performances of LVLMs under different hypergraph representations, we train \texttt{WiseHyGR}, a generalizable router that can choose proper hypergraph representations
for given hypergraph problems. Experimental results validate that \texttt{WiseHyGR} generally enhances the hypergraph
understanding and reasoning abilities of LVLMs, and the benefits can be generalized to downstream out-of-domain
tasks.

The contributions of this work are threefold.

\begin{itemize}\item We construct the \texttt{HyperGVL} benchmark, a new arena to assess LVLMs' abilities on hypergraph understanding and reasoning.

\item We extensively evaluate 12 leading LVLMs on \texttt{HyperGVL} and expose their actual capabilities. The dedicated evaluations from various perspectives contribute 14 valuable observations.

\item Based on the model performance across hypergraph representations, we train  \texttt{WiseHyGR}, a generalizable router to boost LVLMs on hypergraph understanding and reasoning tasks. 
\end{itemize}

\section{The \texttt{HyperGVL} Benchmark}

In this section, we introduce the \texttt{HyperGVL} Benchmark, which is designed to delineate the ability boundaries of LVLMs in handling higher-order structures of hypergraphs.

\subsection{Benchmark Uniqueness}

\textcolor{black}{Table \ref{tab:all_bench} underscores the distinct role of \texttt{HyperGVL} within the landscape of existing benchmarks. Unlike conventional graph-related LVLM benchmarks, \texttt{HyperGVL} delves into higher-order relationships inherent in hypergraphs, surpassing the limitations of ordinary graphs. Additionally, in contrast to text-only hypergraph benchmarks for large language models (LLMs), \texttt{HyperGVL} integrates visual perception effects unique to LVLMs, as well as enhances task diversity and complexity through the inclusion of intricate reasoning challenges.} More related works are introduced in Appendix \ref{app:related}.


\subsection{Hypergraph Organization}
\label{sec:struct}
The hypergraphs in HyperGVL involve meticulous considerations to ensure a reasonable organization. First, the HyperGVL benchmark comprises an equal proportion of synthetic and real-world hypergraphs. Synthetic hypergraphs are generated using both random and regular-structured methods, providing a controlled environment for testing. In contrast, real-world hypergraphs are from anonymized citation and protein networks, offering practical insights into real-world applications.

\textcolor{black}{To ensure balanced complexity for comprehensive evaluation, we employed the scale partition protocol from \citet{llm4hypergraph}, and organized hypergraphs by vertex count into three scale groups: small, medium, and large, with a distribution ratio of 1:2:1.} This categorization facilitates the assessment of model performance across varying levels of complexity. Detailed descriptions of the processes used to obtain these hypergraphs are provided in Appendix \ref{sec:structure}.

\subsection{Benchmark Tasks}
\label{sec:task}
In this section, we introduce the tasks with design considerations in \texttt{HyperGVL}. 

\subsubsection{Design Principle}
\textcolor{black}{The tasks in \texttt{HyperGVL} are designed around two core dimensions: \textbf{assessed capability} and 
\textbf{response type}}.

For \textbf{assessed capabilities}, tasks are divided into two main categories: understanding and reasoning. The understanding tasks evaluate three key atomic abilities: (1) \textit{basic element capture}, which involves recognizing vertices and hyperedges; (2) \textit{adjacency perception}, which entails understanding adjacency relationships among vertices; and (3) \textit{heuristic computation}, which includes computing heuristics such as vertex degree and hyperedge order (i.e., the number of vertices in a hyperedge). On the other hand, reasoning tasks assess model abilities in terms of (1) \textit{algorithms}, which involve solving problems with definitive algorithms, and (2) \textit{planning}, where problems are NP-hard and lack definitive algorithms, requiring models to actively plan and devise valid solutions.

Based on these assessed capabilities, all tasks are categorized into a four-level \textbf{difficulty} hierarchy: Level-1 (querying single atomic capability), Level-2 (combining compound atomic capabilities), Level-3 (polynomial-solvable algorithms), and Level-4 (NP-hard  planning). This stratification aligns with task complexity \cite{bylander1994computational}, and we aim to verify whether it is consistent with the actual capability spectrum of LVLMs.

For \textbf{response types}, tasks are categorized into four types: (1) \textit{counting}, (2) \textit{computing}, (3) \textit{decision}, and (4) \textit{descriptive} tasks. This taxonomy enables a comprehensive evaluation of LVLMs across diverse cognitive processes.

Unlike \texttt{LLM4Hypergraph} \cite{llm4hypergraph}, which presents relatively simple tasks for recent models (e.g., Gemini-3 Flash achieved over 90\% zero-shot accuracy in its \textcolor{black}{13} out of 15 tasks in our testing), the proposed benchmark introduces more challenging tasks that require reasoning beyond structural understanding. This design aligns with evolving model capabilities and \textcolor{black}{leaves ample room for their future improvement}. Overall, the task distribution in \texttt{HyperGVL}  encompasses a wider range of difficulty and diversity, establishing a comprehensive evaluation framework for LVLMs.
\begin{table*}[t]
    \vspace{-0.2cm}
    \vspace{-0.3cm}
    \footnotesize
    \centering
    \resizebox{\textwidth}{!}{
    \begin{tabular}{cccclc}
    \hline
    \toprule
    \textbf{Task} & \textbf{Capability} & \textbf{Response Type} & \textbf{Difficulty} & \makecell[c]{Example} & \#Sample \\ 
    \midrule
 \multicolumn{5}{l}{\textbf{\textit{Understanding Tasks}}} & 42,000\\
    \midrule
        VC & Element & Counting & Level-1 & \makecell[l]{Q: How many vertices are in the hypergraph G?\\A: 15.} & 7,000 \\
        \hline
        HEC & Element & Counting & Level-1 & \makecell[l]{Q: How many vertices are in the hypergraph G?\\A: 23.} & 7,000 \\
        \hline
     Ne & Adjacency & Descriptive & Level-1&\makecell[l]{Q: What are the direct neighbors of vertex v4 in hypergraph G?\\A: v0, v3, v5.} & 7,000 \\
     \hline
     DVC & Heuristic\&Element & Counting & Level-2 & \makecell[l]{Q: How many vertices have degree 3 in hypergraph G?\\A: 7.} & 7,000 \\
     \hline
    OEC & Heuristic\&Element & Counting &Level-2 & \makecell[l]{Q: How many hyperedges have order 4 in hypergraph G?\\A: 8.} & 7,000 \\
    \hline
     ONe & Heuristic\&Adjacency & Descriptive & Level-2&\makecell[l]{Q: What are the neighbors of vertex v5 when only considering\\hyperedges with order >= 2 in hypergraph G?\\A: v0, v3.} & 7,000 \\
     
     \midrule
      \multicolumn{5}{l}{\textbf{\textit{Reasoning Tasks}}} & 42,000\\
    \midrule
     OSP & Algorithm & Computing &Level-3 & \makecell[l]{Q: What is the order-weighted shortest path length from v4 to v8?\\A: 8.} & 7,000 \\
     \hline
     OMF & Algorithm & Computing &Level-3 & \makecell[l]{Q: What is the order-weighted maximum flow from v4 to v8?\\A: 19.} & 7,000 \\
     \hline
     ISM & Algorithm & Decision & Level-3 & \makecell[l]{Q: Are these two hypergraphs isomorphic?\\A: Yes.} & 7,000 \\
     \hline
     3-CL & Planning & Descriptive &\makecell[c]{Level-4\\(NP-hard)} & \makecell[l]{Q: Please provide a 3-coloring strategy such that each hyperedge \\contains nodes with at least 2 different colors.\\A: Coloring:[v0:c0,v1:c1,v2:c2,v3:c0,v4:c1,v5:c2].} & 7,000 \\
     \hline
     SHC & Planning & Descriptive &\makecell[c]{Level-4\\(NP-hard)} & \makecell[l]{Q: Please identify a strict hypercycle in the hypergraph G. \\A: Cycle:[e0,e3,e2,e4,e1].} & 7,000 \\
     \hline
     HHM & Planning & Descriptive &\makecell[c]{Level-4\\(NP-hard)} & \makecell[l]{Q: Please provide a valid Hamiltonian path from v1 to v0.\\(Hamiltonian path = path visiting all vertices exactly once). \\A: Path:[e0,e1,e2,e3].} & 7,000 \\
    \bottomrule
    \hline\\
    \end{tabular}}
    \vspace{-0.5cm}
    \caption{Properties, statistics, and examples of all hypergraph understanding and reasoning tasks in \texttt{HyperGVL}.}
\label{tab:task}
\vspace{-15pt}
\end{table*}

\subsubsection{Task Descriptions}
All tasks in \texttt{HyperGVL} are introduced briefly in this section. More details are in Appendix \ref{app:task}.

\textbf{Hypergraph understanding tasks}
are designed to evaluate in terms of the composition, topology, and fundamental heuristics of hypergraphs. Those tasks are mainly categorized into six types as follows.
\begin{itemize}
\item \textit{Vertex Counting (VC)}: Counting the number of vertices in a given hypergraph.
\item \textit{Hyperedge Counting (HEC)}: Counting the number of hyperedges in a given hypergraph.
\item \textit{Neighbors (Ne)}: Identifying direct neighbors of a specified vertex connected by hyperedges.
\item \textit{Degree-specified Vertex Counting (DVC)}: Counting vertices with a specific degree value in the hypergraph.
\item \textit{Order-specified Hyperedge Counting (OEC)}: Counting hyperedges with a specific order in the hypergraph.
\item \textit{Order-filtered Neighbors (ONe)}: Identifying neighbors of a vertex when only considering hyperedges with their orders no smaller than a specified threshold.
\end{itemize}
The associated \textbf{assessed capabilities}, \textbf{difficulty levels}, and \textbf{response types} of those tasks are detailed at the top of Tab.~\ref{tab:task}, along with examples.

\textbf{Hypergraph reasoning tasks} are designed to tackle complex, multi-step inferential challenges within hypergraphs. Beyond understanding hypergraph structures and computing heuristics, these tasks require organizing the atom capabilities into a sophisticated iterative process to tackle complex hypergraph problems. Those tasks are mainly classified into six types as follows.
\begin{itemize}
    \item \textit{Order-weighted Shortest Path (OSP)}: Computing the shortest path length between two vertices, where the hyperedge order serves as the distances.
    \item \textit{Order-weighted Maximum Flow (OMF)}: Computing the maximum flow between two vertices, where the hyperedge order determines the edge capacity.
    \item \textit{Isomorphism Recognition (ISM)}: Determining whether two hypergraphs are isomorphic.
    \item \textit{Hypergraph 3-Coloring (3-CL)}: Providing a valid 3-coloring, where each hyperedge contains at least two distinct colors.
    \item \textit{Strict Hypercycle (SHC)}: Searching a strict hypercycle in the hypergraph, where adjacent hyperedges share exactly one vertex.
    \item \textit{Hypergraph Hamilton Path (HHM)}: Planning a Hamiltonian path, which visits all the vertices in the hypergraph exactly once,  with a given vertex as the starting point of the path and another one as the ending point.
\end{itemize}
The associated \textbf{assessed capabilities}, \textbf{difficulty levels}, and \textbf{response types} of those tasks are detailed in the bottom of Tab.~\ref{tab:task}.

\subsection{Hypergraph Representations}
\label{sec:intro_rep}

\begin{figure*}
    \centering
    \vspace{-0.5cm}
    \includegraphics[width=\linewidth]{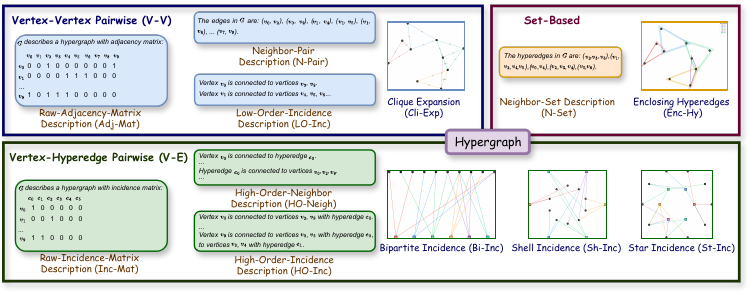}
    \caption{The 7 textual hypergraph representations and 5 visual hypergraph representations in \texttt{HyperGVL}.}
    \label{fig:language}
    \vspace{-10pt}
\end{figure*}

Hypergraph representations are crucial for evaluating the capabilities of LVLMs within hypergraphs, as distinct representations introduce unique perceptual biases \cite{wei2024gita, llm4hypergraph}. Unlike LLMs that rely solely on text, LVLMs benefit from a synergistic perception of both visual and textual information. Therefore, testing LVLMs on \texttt{HyperGVL} should not only consider textual or visual hypergraph representations individually, but also study their combinations. To comprehensively assess LVLMs across different representations, \texttt{HyperGVL} utilizes seven textual representations and five visual representations, resulting in 35 combinations of hypergraph representations (7 textual $\times$ 5 visual). To clarify, the 84,000 vision-language QA samples in \texttt{HyperGVL} are related to 2,400 meta problems (each task 200), where each meta problem $p$ corresponds to 35 vision-language QA samples with varying hypergraph representations. 

In the following, we introduce these hypergraph representations, and provide examples in Fig. \ref{fig:language}.

\paragraph{Textual Hypergraph Representations.} 
By following \cite{llm4hypergraph},
we use the following 7 textual hypergraph representations in \texttt{HyperGVL}. Their prompts are demonstrated in Appendix \ref{app:prompt_t}: 
\begin{itemize}
    \item \textit{Low-Order Incidence Description (LO-Inc)}: Listing pairwise vertex connections, e.g., ``Vertex $v_1$ is linked to vertices $v_2$ and $v_3$''.
    \item \textit{Neighbor-Pair Description (N-Pair)}: Enumerating pairs of vertices sharing a hyperedge, e.g., ``$(v_1, v_2), (v_1, v_3)$''.
    \item \textit{Raw Adjacency Matrix Description (Adj-Mat)}: Employing a matrix where the binary entry indicates the connection between vertex pairs.
    \item \textit{High-Order Neighbor Description (HO-Neigh)}: Detailing neighborhood information in two parts, outlining vertex-to-hyperedge and hyperedge-to-vertex connections.
    \item \textit{High-Order Incidence Description (HO-Inc)}: Incorporating higher-order relationships into LO-Inc, e.g., ``Vertex $v_1$ is connected to vertices $v_2$ and $v_3$ via hyperedge $e_1$''.
    \item \textit{Neighbor-Set Description (N-Set)}: Listing complete vertex sets associated with each hyperedge, e.g., ``$(v_1, v_2, v_3)$''.
    \item \textit{Raw Incidence Matrix Description (Inc-Mat)}: Uses a matrix where the binary entry signifies the inclusion of a vertex in a hyperedge.
\end{itemize}

\paragraph{Visual Hypergraph Representations.} For the vision branch, \texttt{HyperGVL} incorporates five visual hypergraph representations as follows, with details of their generations in Appendix \ref{app:visual} and examples in Fig.~\ref{fig:language}.
\begin{itemize}
\item \textit{Enclosing Hyperedges (Enc-Hy)}: Visualizing hyperedges as geometric enclosures, with vertices inside enclosures indicating membership in the corresponding hyperedge. 
\item \textit{Bipartite Incidence (Bi-Inc)}: Modeling hypergraphs as bipartite structures, with vertices and hyperedges as two disjoint sets arranged in separate layers. Straight lines between layers depict vertex-hyperedge membership.
\item \textit{Shell Incidence (Sh-Inc)}: Similar to the \emph{Bipartite Incidence} representation, but arranging vertices on an outer circle and places hyperedges internally, with radial connections depicting the membership.  
\item \textit{Star Incidence (St-Inc)}: Similar to Bipartite Incidence, but treating each hyperedge as a star center connected to its member vertices.
\item \textit{Clique Expansion (Cli-Exp)}: Transforming hyperedges into vertex-vertex pairwise connections, with hyperedge IDs labeled to preserve hypergraph semantics.  
\end{itemize}
Based on the above encoding approaches, those textual and visual hypergraph representations are categorized into three representation categories: \textbf{(1) Vertex-Vertex Pairwise (V-V)} category, which includes \textit{LO-Inc, N-Pair, Adj-Mat}, and \textit{Cli-Exp}, focuses on vertex-vertex neighborships with hyperedge semantics implied through identifiers; \textbf{(2) Vertex-Hyperedge Pairwise (V-E)} category, comprising \textit{HO-Neigh, HO-Inc, Inc-Mat, Bi-Inc, Sh-Inc}, and \textit{St-Inc}, explicitly decomposes high-order relationships into pairwise mappings from vertices to their associated hyperedges; \textbf{(3) Set-Based} category, featuring \textit{N-Set} and \textit{Enc-Hy}, treats hyperedges as holistic units, introducing their constituent vertices together.



\section{Benchmarking LVLMs on \texttt{HyperGVL}}
\begin{table*}[t]
\vspace{-0.03cm}
   \centering
    \resizebox{\linewidth}{!}{
    \renewcommand\arraystretch{1.2}
\begin{tabular}{lccccccccccccccc}
\toprule
\multicolumn{1}{l}{\multirow{2.5}{*}{\textbf{Models}}}   &  \multicolumn{7}{c}{\textbf{Understanding}}   & \multicolumn{7}{c}{\textbf{Reasoning}}                                                  \\
\cmidrule(lr){2-8} \cmidrule(lr){9-15}
\multicolumn{1}{c}{}                   & \textbf{VC}   & \textbf{HEC}    & \textbf{Ne}    & \textbf{DVC} & \textbf{OEC} & \textbf{ONe} & \textbf{Avg.U} & \textbf{OSP} & \textbf{OMF} & \textbf{ISM} & \textbf{3-CL} & \textbf{SHC} & \textbf{HHM} & \textbf{Avg.R} \\
\midrule
\cellcolor{yellow!20}LLaVA-1.5-7B
& \textcolor{blue}{14.10} & \textcolor{blue}{\enspace8.63} & \textcolor{blue}{\enspace2.64} & 10.74 & 16.57 & \textcolor{blue}{\enspace1.61} & \textcolor{blue}{\enspace9.05} & \enspace5.57 & \enspace4.51 & 44.53 & 10.04 & \enspace8.44 & \enspace0.57 & 12.28 \\
\cellcolor{yellow!20}LLaVA-Next-7B
& 19.56 & 11.17 & \enspace6.07 & \textcolor{blue}{\enspace8.83} & 14.59 & \enspace8.39 & 11.44 & \enspace6.83 & \enspace3.84 & 42.27 & 15.84 & \enspace2.83 & \enspace0.31 & 11.99 \\
\cellcolor{yellow!20}Molmo-7B-D-0924
& 42.24 & 17.94 & \enspace7.96 & 13.24 & 15.81 & 14.40 & 18.60 & 10.53 & \enspace6.74 & 45.90 & \enspace6.63 & \enspace6.80 & \enspace0.40 & 12.83 \\
\cellcolor{yellow!20}InternVL2.5-8B
& 73.34 & 22.93 & 10.73 & 15.93 & 18.26 & 22.00 & 27.20 & 10.94 & \enspace6.70 & 51.06 & 22.76 & \enspace5.94 & \enspace0.69 & 16.35 \\
\cellcolor{yellow!20}GLM-4V-9B
& 91.34 & 39.37 & 15.46 & 17.89 & 23.33 & 18.90 & 34.38 & 13.51 & \enspace8.89 & 55.54 & 20.57 & 14.19 & \enspace0.67 & 18.90 \\
\cellcolor{yellow!20}InternVL3-8B
& 88.93 & 47.93 & 39.17 & 27.47 & 35.71 & 30.54 & 44.96 & 16.23 & 11.69 & 50.54 & 38.04 & \enspace5.24 & \enspace1.41 & 20.52 \\
\cellcolor{yellow!20}Qwen2.5-VL-7B
& 95.41 & 41.94 & 36.19 & 24.63 & 29.01 & 27.17 & 42.39 & 15.73 & 14.53 & 59.23 & 26.31 & 10.51 & \enspace1.50 & 21.30 \\
\cellcolor{yellow!20}Qwen3-VL-8B & 99.60 & 59.41 & 55.20 & 52.34 & 49.14 & 43.13 & 59.80 & 32.10 & 25.09 & 42.79 & 57.07 & 46.57 & \enspace2.93 & 34.43 \\
\cellcolor{yellow!20}Gemma3-12B & \textcolor{red}{99.80} & 57.91 & 73.34 & 51.92 & 47.35 & 50.87 & 63.53 & 29.41 & 22.71 & 51.29 & 33.69 & 13.39 & \enspace2.67 & 25.53 \\
\cellcolor{yellow!20}Deepseek-VL2-27B & 48.59 & 17.73 & 2.79 & 9.81 & \textcolor{blue}{10.60} & 5.11 & 15.77 & \textcolor{blue}{\enspace3.54} & \textcolor{blue}{\enspace3.07} & \textcolor{blue}{40.76} & \textcolor{blue}{\enspace1.40} & \textcolor{blue}{\enspace0.43} & \textcolor{blue}{\enspace0.03} & \textcolor{blue}{\enspace8.21}\\ 
\cellcolor{orange!20}GPT-4o
& 98.29 & 80.29 & 83.43 & 56.57 & 50.57 & 82.86 & 66.91 & 39.43 & 26.29 & 59.14 & 56.86 & 25.14 & \enspace3.13 & 35.00\\
\cellcolor{orange!20}Gemini-3 Flash
& 99.73 & \textcolor{red}{98.31} & \textcolor{red}{88.20} & \textcolor{red}{77.11} & \textcolor{red}{77.83} & \textcolor{red}{97.13} & \textcolor{red}{89.72} & \textcolor{red}{82.09} & \textcolor{red}{76.29} & \textcolor{red}{64.60} & \textcolor{red}{79.34} & \textcolor{red}{66.44} & \textcolor{red}{\enspace3.50} & \textcolor{red}{62.04} \\
\midrule
\cellcolor{gray!20}  Average & 72.58 & 41.96 & 37.71 & 30.54 & 32.40 & 30.90 & 40.31 & 22.16 & 17.53 & 50.64 & 30.71 & 17.16 & 1.48  & 23.28 \\
\bottomrule 
\end{tabular}}
 \vspace{-0.15cm}
        \caption{Model performance (Acc) on \texttt{HyperGVL} benchmark, where \textbf{Avg.U} and \textbf{Avg.R} represent the average accuracy across understanding and reasoning tasks, respectively. Open- and closed-source LVLMs are colored in \colorbox{yellow!20}{.}  and \colorbox{orange!20}{.} backgrounds.  The best and worst results are colored in \textcolor{red}{red} and \textcolor{blue}{blue}, respectively.}
    \label{tab:QA_results}
    \vspace{-15pt}
\end{table*}
\subsection{Experimental Setup}
We conduct evaluations 
on 12 leading LVLMs, including open-source LVLMs: LLaVA1.5-7B \cite{llava1.5}, LLaVA-Next-7B \cite{liu2024llavanext}, Molmo-7B-D-0924 \cite{deitke2025molmo}, InternVL2.5-8B \cite{internvl2.5}, InternVL3-8B \cite{zhu2025internvl3}, GLM-4V-9B \cite{glm2024chatglm}, Qwen2.5-VL-7B \cite{qwen25vl}, Qwen3-VL-8B \cite{Qwen3-VL}, Gemma3-12B \cite{team2025gemma}, and Deepseek-VL2-27B \cite{deepseekvl2} as well as GPT-4o \cite{gpt4o} and Gemini-3 Flash \cite{gemini3}, which are strong closed-source LVLMs.
For all LVLMs, the \textbf{zero-shot} setting is adopted for the evaluation. The LVLMs are required to provide answers in predefined formats, but the factually accurate answers with incorrect formats are also considered to be correct after manual review. Prompts used in the evaluation are detailed in Appendix \ref{app:prompt_task}.  
More details of experimental settings are in Appendix \ref{app:evaluation}.

\subsection{Main Results}
Tab.~\ref{tab:QA_results} shows the performance of various LVLMs on \texttt{HyperGVL}, with results averaged across all hypergraph representation combinations. Key observations include:

\noindent{\textbf{\textit{Observation 1}}}: Basic hypergraph components and neighborhood understanding (Level-1 VC, HEC, and Ne) could be challenging for open-source LVLMs. Advanced open-source LVLMs like Qwen3-VL and Gemma3 excel in vertex counting (VC), but they are still unsatisfactory in capturing hyperedges (HEC) and neighborhoods (Ne). In contrast, closed-source LVLMs perform robustly across these foundational capabilities.

\noindent{\textbf{\textit{Observation 2}}}: Level-2 tasks (i.e., DVC, OEC, ONe) introduce complexity beyond Level-1 tasks by incorporating heuristic computations as constraints, leading to lower performance in DVC, OEC, and ONe than VC, HEC, and Ne as expected.

\noindent{\textbf{\textit{Observation 3}}}: Reasoning tasks pose significant challenges for all LVLMs, with Avg.R across models (23.28\%) notably lower than Avg.U (40.31\%). Even the most powerful LVLM, Gemini-3 Flash, achieves only 62.04\% in terms of \textit{Avg.R}, while most open-source LVLMs score below 25\%. 

\noindent{\textbf{\textit{Observation 4}}}: The hypergraph reasoning performance of LVLMs generally aligns with task complexity. Overall, Level-4 NP-hard tasks (i.e., 3-CL, SHC, HHM) are more challenging than Level-3 tasks, with HHM as the most intractable, that is, no model exceeds 6\% in terms of the accuracy on this task. This demonstrates that planning constitutes a complex capability that requires substantial improvement for current LVLMs. 3-CL is a bit easier due to its abundance of valid solutions, which can be derived via local constraints. In contrast, SHC and HHM require global consistency, feature far fewer valid solutions, and are highly susceptible to intermediate inference errors, rendering them significantly challenging for LVLMs. 

\noindent{\textbf{\textit{Observation 5}}}: All LVLMs perform well in ISM, which may attributes to the visual representations directly reveals similar patterns among isomorphic hypergraphs (illustrated in Appendix \ref{app:ism}). 

\noindent{\textbf{\textit{Observation 6}}}: Closed-source LVLMs significantly outperform open-source counterparts in both understanding and reasoning tasks, with the highest Avg.U reaching 89.72\% for closed-source LVLMs (Gemini-3 Flash) compared to 63.53\% for open-source LVLMs (Gemma3). For reasoning tasks, the disparity is 62.04\% (Gemini-3 Flash) vs. 34.43\% (Qwen3-VL) in terms of Avg.R. This highlights substantial room for improvement in the hypergraph understanding and reasoning capabilities of open-source LVLMs.

\noindent{\textbf{\textit{Observation 7}}}: Parameter scale does not completely determine the performance of LVLMs. Smaller models are not always worse than larger ones. For example, Qwen3-VL-8B performs better at reasoning tasks than Gemma3-12B and Deepseek-VL2-27B. Moreover, there are significant differences even among models with comparable sizes (e.g., LLaVA-1.5-7B vs. Qwen3-VL-8B), suggesting that enhancing LVLM's hypergraph capabilities requires specialized approaches rather than merely increasing the model size.

\begin{table*}[h!tbp]
   \centering
    \resizebox{\linewidth}{!}{
    \renewcommand\arraystretch{1.2}
\begin{tabular}{lcccccccccccccccc}
\toprule
\multicolumn{2}{l}{\multirow{2.5}{*}{\textbf{Representations}}}   &  \multicolumn{7}{c}{\textbf{Understanding}}   & \multicolumn{7}{c}{\textbf{Reasoning}}                                                  \\
\cmidrule(lr){3-9} \cmidrule(lr){10-16}
\multicolumn{2}{c}{}                   & \textbf{VC}   & \textbf{HEC}    & \textbf{Ne}    & \textbf{DVC} & \textbf{OEC} & \textbf{ONe} & \textbf{Avg.U} & \textbf{OSP} & \textbf{OMF} & \textbf{ISM} & \textbf{3-CL} & \textbf{SHC} & \textbf{HHM} & \textbf{Avg.R}
\\
\midrule
\multirow{5}{*}{\rotatebox{90}{V-V}} &  \cellcolor{yellow!20} LO-Inc & 75.02 & 23.45 & 40.53 & \textcolor{blue}{17.19} & \textcolor{blue}{16.94} & \textcolor{red}{50.12} & 36.51 & 13.86 & 11.71 & 50.34 & 24.11 & 10.96 & 0.79 & 18.64 \\
& \cellcolor{yellow!20} N-Pair & 75.00 & \textcolor{blue}{14.90} & 32.92 & 18.52 & 17.27 & 33.64 & 31.34 & 17.15 & 13.43 & 50.54 & \textcolor{blue}{19.12} & 15.26 & 0.80 & 19.40 \\
& \cellcolor{yellow!20} Adj-Mat & \textcolor{red}{76.28} & 21.79 & \textcolor{blue}{25.81} & 18.87 & 18.38 & 18.94 & \textcolor{blue}{29.31} & \textcolor{blue}{13.74} & \textcolor{blue}{10.18} & 50.53 & 24.80 & \textcolor{blue}{10.11} & \textcolor{blue}{0.78} & \textcolor{blue}{18.37} \\
& \cellcolor{orange!20} Cli-Exp & 70.39 & 40.04 & 38.00 & 31.31 & 32.29 & 31.02 & 39.81 & 22.98 & 17.76 & \textcolor{red}{56.17} & 31.15 & 18.06 & 1.55 & 24.62 \\
& \cellcolor{gray!20} V-V Avg. & 74.17 & 25.05 & 34.32 & 21.47 & 21.22 & 33.43 & 34.24 & 16.93 & 13.27 & 51.90 & 24.80 & 13.60 & 0.98 & 20.26 \\
\midrule
\multirow{7}{*}{\rotatebox{90}{V-E}} &  \cellcolor{yellow!20} HO-Neigh & 73.28 & \textcolor{red}{67.03} & 44.20 & \textcolor{red}{48.65} & 53.19 & 30.43 & \textcolor{red}{52.10} & 30.45 & 23.84 & 51.72 & 37.29 & 21.48 & 2.03 & 27.81 \\
& \cellcolor{yellow!20} HO-Inc & 73.82 & 61.37 & 46.16 & 38.53 & 40.36 & 40.06 & 49.35 & 27.44 & 21.68 & 51.05 & \textcolor{red}{39.95} & \textcolor{red}{22.59} & \textcolor{red}{2.51} & 27.55 \\
& \cellcolor{yellow!20} Inc-Mat & \textcolor{blue}{63.80} & 40.42 & 27.78 & 36.59 & 26.64 & \textcolor{blue}{10.05} & 33.51 & 19.80 & 17.15 & 50.05 & 30.29 & 16.65 & 1.58 & 22.60 \\
& \cellcolor{orange!20} Bi-Inc & 70.25 & 40.95 & 38.08 & 30.21 & 32.56 & 31.02 & 39.81 & 21.67 & 17.49 & 47.36 & 30.43 & 16.58 & 1.51 & 22.52 \\
& \cellcolor{orange!20} Sh-Inc & 74.43 & 41.57 & 37.20 & 29.73 & 31.63 & 30.74 & 40.18 & 21.88 & 17.28 & 47.58 & 29.77 & 16.30 & 1.39 & 22.38 \\
& \cellcolor{orange!20} St-Inc & 74.72 & 42.00 & 37.32 & 30.36 & 32.04 & 30.45 & 40.45 & 22.03 & 17.37 & \textcolor{blue}{46.96} & 30.69 & 16.43 & 1.41 & 22.30 \\
& \cellcolor{gray!20} V-E Avg. & 71.72 & 48.89 & 38.46 & 35.68 & 36.07 & 28.79 & 42.57 & 23.88 & 19.14 & 49.12 & 33.07 & 18.34 & 1.74 & 24.19 \\
\midrule
\multirow{3}{*}{\rotatebox{90}{Set-Based}} &  \cellcolor{yellow!20} N-Set & 70.89 & 64.74 & 46.55 & 35.40 & \textcolor{red}{54.02} & 33.09 & 50.08 & \textcolor{red}{32.70} & \textcolor{red}{24.72} & 49.67 & 39.65 & 21.92 & 1.85 & \textcolor{red}{28.44} \\
& \cellcolor{orange!20} Enc-Hy & 73.11 & 45.23 & 37.92 & 31.07 & 33.48 & 31.29 & 41.32 & 22.26 & 17.76 & 55.72 & 31.26 & 19.61 & 1.53 & 24.70 \\
& \cellcolor{gray!20} Set-Based Avg. & 72.00 & 54.99 & 42.24 & 33.24 & 43.75 & 32.19 & 45.70 & 27.48 & 21.24 & 52.70 & 35.46 & 20.77 & 1.69 & 26.57 \\
\bottomrule
\end{tabular}}
 \vspace{-0.2cm}
       \caption{Average performance (Acc) with different hypergraph representations. Textual and visual representations are distinguished by \colorbox{yellow!20}{.}  and \colorbox{orange!20}{.} backgrounds. The best and worst performance are colored in \textcolor{red}{red} and \textcolor{blue}{blue}.}
    \label{tab:hyper_rep}
    \vspace{-10pt}
\end{table*}

\subsection{Comparison of LVLMs and LLMs}
We compare the performance of two representative LVLMs, GLM-4V and InternVL3, with their corresponding LLMs, GLM-4 and InternLM3.

\begin{figure}[t]
    \centering
    \includegraphics[width=0.45\textwidth]{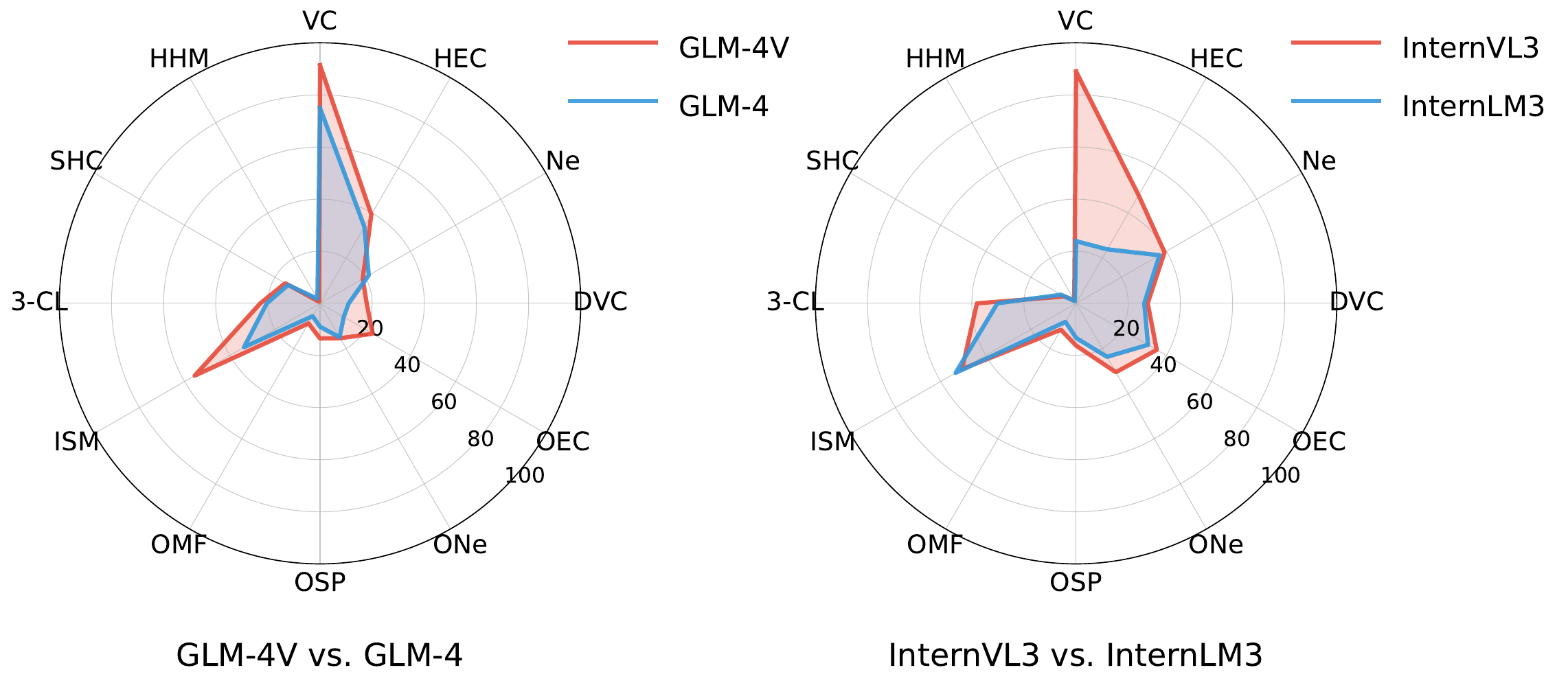}
    \caption{Comparison between the capability boundaries of LVLMs (GLM-4V \& InternVL3) and corresponding LLMs (GLM-4 \& InterLM3).}
    \label{fig:text_results}
    \vspace{-10pt}
\end{figure}

\noindent{\textbf{\textit{Observation 8}}}: As illustrated in Fig.~\ref{fig:text_results}, both LVLMs consistently outperform their LLMs on all understanding and reasoning tasks. 
This demonstrates the advantages of visual-textual synergy in LVLMs over text-only processing in LLMs on hypergraphs. These results further underscore the imperative of integrating LVLMs into hypergraph scenarios, a step that \texttt{HyperGVL} has initiated.

\subsection{Results on Hypergraph Representations}
\label{sec:representations}

Table \ref{tab:hyper_rep} presents the average performance across various hypergraph representations. The results for each textual representation are averaged over all LVLMs and visual representations, while the results for each visual representation are averaged over all LVLMs and textual representations.
Key observations include:

\noindent{\textbf{\textit{Observation 9}}}: Among textual representations, `N-Set' demonstrates superior average performance in both understanding and reasoning tasks. Similarly, among visual representations, `Enc-Hy' excels in both areas. Both representations are set-based, underscoring the overall importance of their holistic hyperedge expression.


\noindent{\textbf{\textit{Observation 10}}}: Representation-induced variability differs substantially across modalities. \textbf{Textual} encodings yield a much wider performance range (higher ceiling and lower floor), while \textbf{visual} encodings lead to comparatively more moderate variations, indicating that LVLMs are more representation-sensitive in the text branch and more representation-robust in the visual branch.

\noindent{\textbf{\textit{Observation 11}}}: The V--V pairwise encodings are comparatively more suitable for vertex-level primitives (e.g., VC and neighbor-related queries), but they consistently lag behind when tasks require explicit hyperedge semantics (e.g., hyperedge cardinality/order and planning with global constraints). This suggests that LVLMs remain limited in recovering high-order hyperedge information from purely vertex--vertex adjacency.
\begin{figure}[t]
    \centering
    \includegraphics[width=0.45\textwidth]{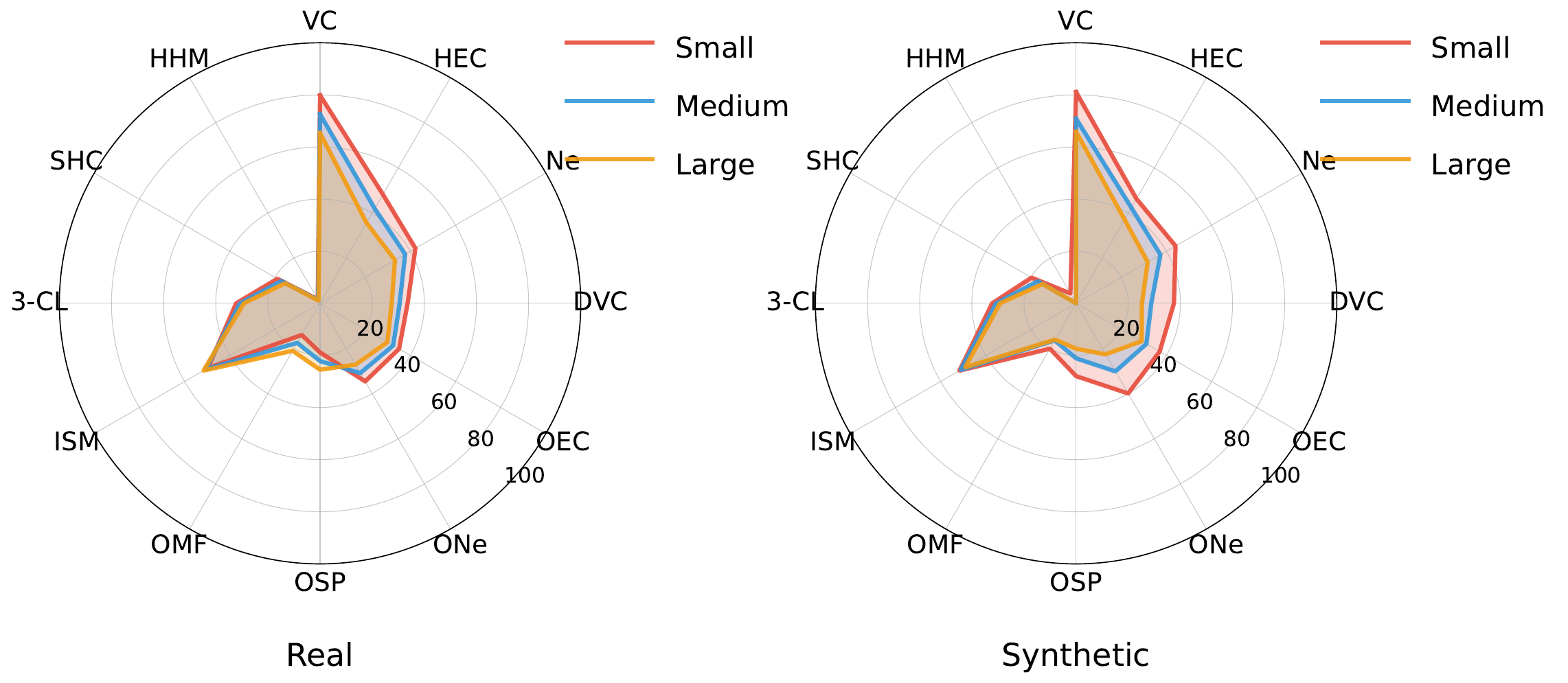}
    \caption{Performance for real-world/synthetic
    hypergraphs at different sizes.}
    \label{fig:structure}
    \vspace{-10pt}
\end{figure}

\noindent{\textbf{\textit{Observation 12}}}: \emph{HO-Inc} is the top-performing representation on planning tasks (e.g., 3-CL, SHC, HHM). We attribute this to its explicit vertex--hyperedge--vertex modeling with hyperedge IDs (e.g., ``$v_1$ is connected to $v_2, v_3$ via hyperedge $e_1$''), which supports step-wise navigation and constraint checking during multi-step solution construction.

\noindent{\textbf{\textit{Observation 13}}}: Within textual representations, matrix-style encodings (i.e., Adj-Mat and Inc-Mat) are generally \emph{less competitive} than list/set-style encodings within the same family, likely due to the sparsity  and indexing burdens of matrix representations  hinder reliable extraction of vertex--hyperedge memberships.

\subsection{Impact of Hypergraph Scale}
Fig. \ref{fig:structure} shows the performance of real-world and synthetic hypergraphs at different sizes. We have the following observation:

\noindent{\textbf{\textit{Observation 14}}}: As the hypergraph size increases, the performance on synthetic hypergraphs always deteriorates. This trend is more pronounced for understanding tasks. For real-world hypergraphs, OSP and OMF have better performance on the larger hypergraphs, showing benefits from their more regular connectivity structures.

\section{WiseHyGR Router}
\label{sec:router}
Inspired by our findings on the critical impacts of hypergraph representations, we develop \texttt{WiseHyGR}, a generalizable router aimed at enhancing the LVLM performance on hypergraph understanding and reasoning tasks by intelligently selecting suitable hypergraph representations.

First, we construct a Problem-Representation Mapping (PRM) dataset
$\{(p, R_p)\}$,
where $R_p$ is the representation combination with the highest average accuracy for each vision-language QA set related to the meta problem $p$ in \texttt{HyperGVL}.
When multiple combinations yield the same accuracy, we include multiple $(p, R_p)$ pairs. 

Treating as a classification task, we use the PRM dataset to train a DeBERTaV3-base \cite{he2021debertav3} router named \texttt{WiseHyGR}, with the training and validating split in a ratio of 8:2. The hypergraph in meta problem $p$ is converted to the HO-Neigh representation, chosen for its superior performance in understanding tasks according to Tab.~\ref{tab:hyper_rep}. This hypergraph HO-Neigh description, along with the problem question, serves as input to \texttt{WiseHyGR}, with the label being the optimal representation combination.

We evaluate \texttt{WiseHyGR} based on an open-source 
LVLM (i.e., Qwen3-VL-8B) and a closed-source LVLM (i.e., GPT-4o).
Baselines include: (1) \textbf{\textit{Rand}}, which randomly samples one of the 35 representation combinations, simulating the unguided choice; and (2) \textbf{\textit{Top}}, which uses the most effective textual and visual representations according to Tab.~\ref{tab:hyper_rep} (i.e., N-Set and Enc-Hy). The evaluation involves two settings.

\begin{figure}[t]
    \centering
    \includegraphics[width=0.4\textwidth]{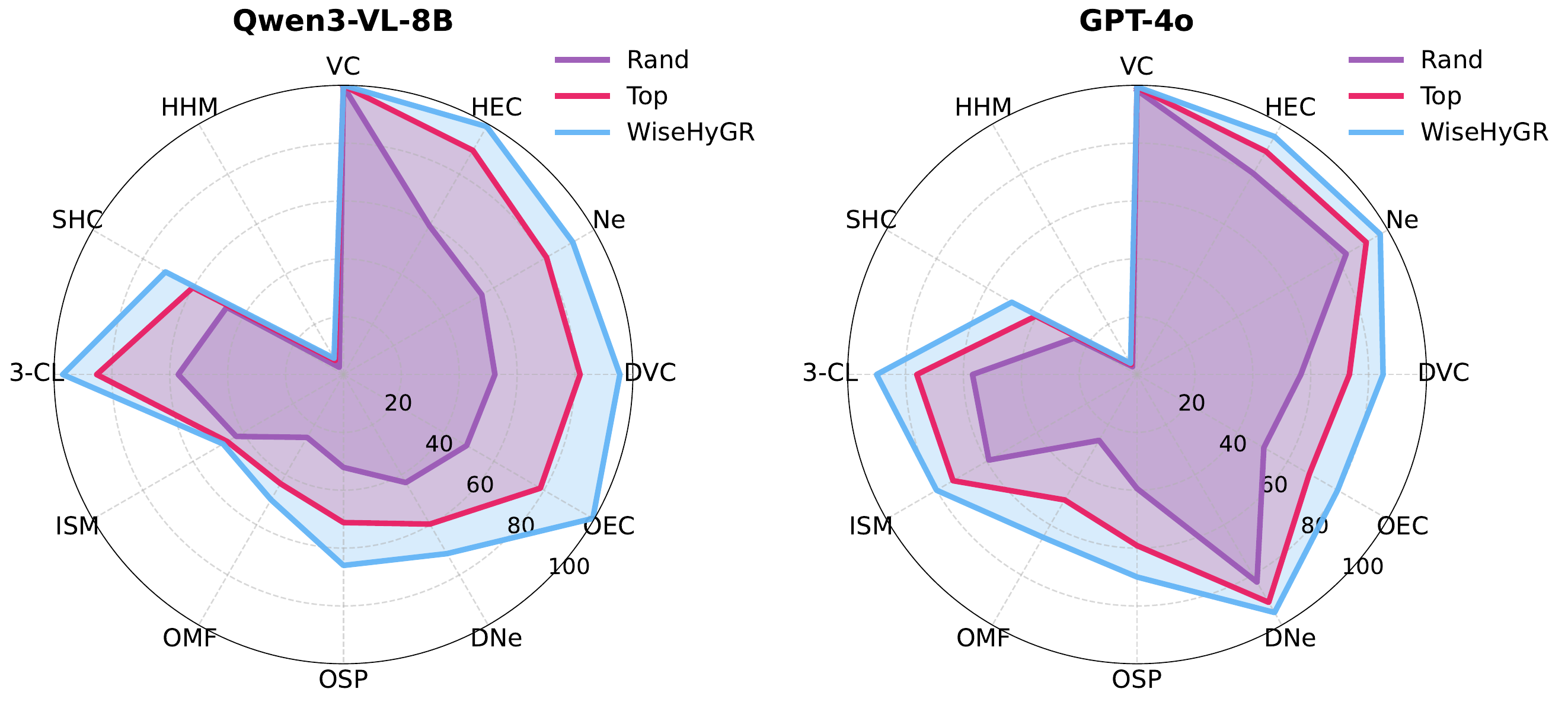}
    \vspace{-5pt}
    \caption{In-domain hypergraph capabilities of LVLMs with WiseHyGR.}
    \label{fig:router_in}
    \vspace{-10pt}
\end{figure}

\paragraph{(1) In-domain hypergraph capabilities}: We generate 1,000 problems per task to assess improvements by \texttt{WiseHyGR} in hypergraph understanding and reasoning. As shown in Fig.~\ref{fig:router_in}, representations chosen by \texttt{WiseHyGR} consistently outperform the baselines across all tasks on both LVLMs, demonstrating its effectiveness in enhancing LVLM capabilities in hypergraph tasks.

\begin{figure}[t]
    \centering
    \includegraphics[width=0.4\textwidth]{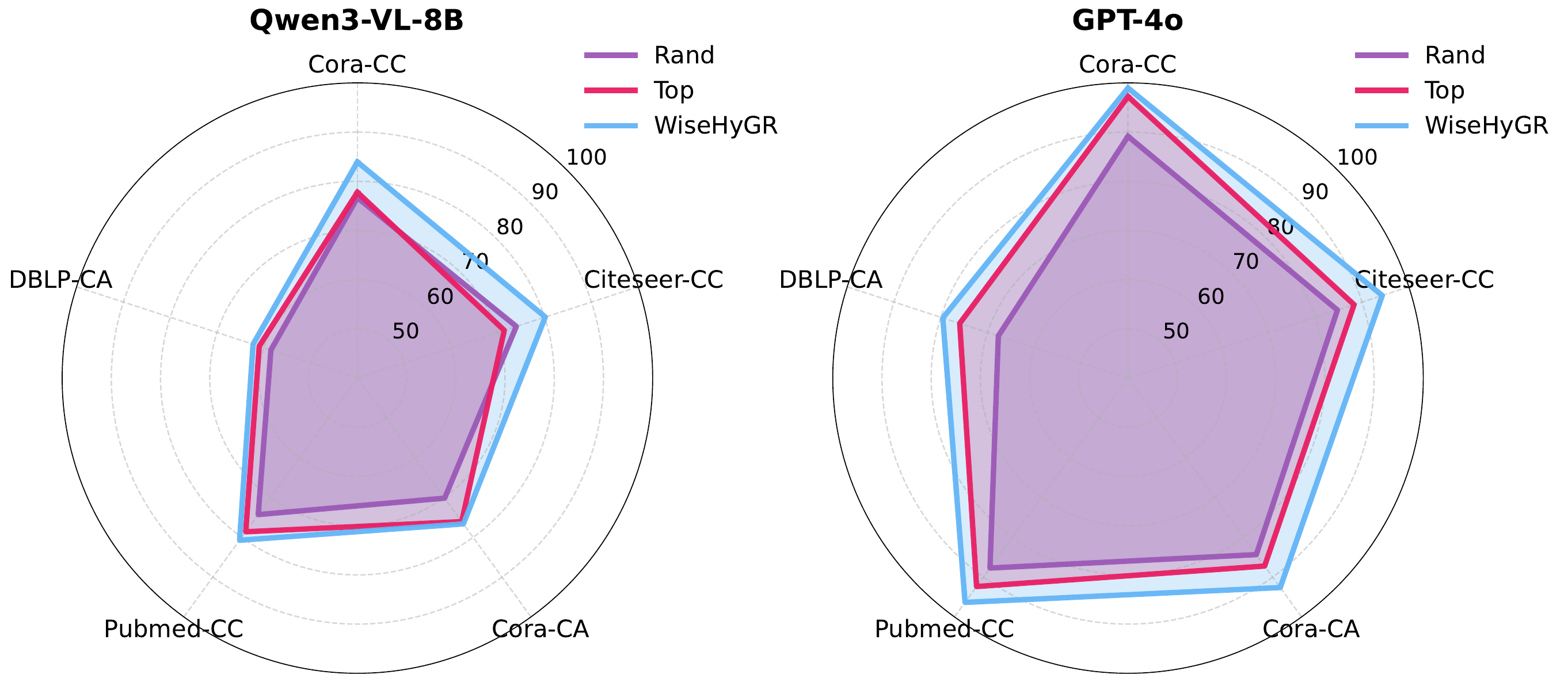}
    \vspace{-5pt}
    \caption{Out-of-domain hypergraph node classification of LVLMs with WiseHyGR.}
    \label{fig:router_out}
    \vspace{-15pt}
\end{figure}

\paragraph{(2) Out-of-domain hypergraph applications}: We further evaluate LVLMs on hypergraph node classification, where \texttt{WiseHyGR} serves as a plug-and-play enhancer. Details of the experimental setting and datas are in Appendix \ref{app:real}. As shown in Fig.~\ref{fig:router_out}, despite being trained solely on \texttt{HyperGVL} tasks and having no exposure to hypergraph node classification tasks, \texttt{WiseHyGR} consistently improves the zero-shot performance of both LVLMs. This shows that \texttt{WiseHyGR} captures the hidden representation preferences embedded in the semantic requirements of tasks with generalizability.


\section{Conclusion}

We present \texttt{Hyper-GVL}, the pioneering benchmark for integrating LVLMs into hypergraph scenarios. 
\texttt{Hyper-GVL} includes 84,000 vision-language QA pairs on synthetic and real-world hypergraphs, spanning 12 tasks that test various capabilities and difficulty levels. 
Our comprehensive evaluation of 12 leading LVLMs provides insightful findings. Additionally, we investigate the impact of diverse textual and visual hypergraph representations and introduce \texttt{WiseHyGR}, a plug-and-play router that enhances LVLMs in hypergraph with generalizability. This work pushes the boundary of LVLMs, marking their initial foray into hypergraph processing.

\section*{Limitations}

While \texttt{HyperGVL} encompasses 12 representative hypergraph tasks to establish robust capability benchmarks, it naturally cannot cover all hypergraph response types, specifically some domain-specific customized tasks,   due to our focus on a general usable benchmark. 

Additionally, resource limitations prevent us from evaluating extremely large-scale models such as Qwen3-VL-235B \cite{Qwen3-VL} and InternVL3-78B \cite{zhu2025internvl3}, despite their potentially greater capabilities on hypergraph understanding and reasoning.



\bibliography{custom}

@inproceedings{llasa,
  title={LLaSA: Large language and structured data assistant},
  author={Xu, Yao and He, Shizhu and Chen, Jiabei and ZengXiangrong, ZengXiangrong and Wang, Bingning and Liu, Guang and Zhao, Jun and Liu, Kang},
  booktitle={Proceedings of the 2025 Conference of the Nations of the Americas Chapter of the Association for Computational Linguistics: Human Language Technologies (Volume 1: Long Papers)},
  pages={1935--1946},
  year={2025}
}

@article{contisciani2022inference,
  title={Inference of hyperedges and overlapping communities in hypergraphs},
  author={Contisciani, Martina and Battiston, Federico and De Bacco, Caterina},
  journal={Nature communications},
  volume={13},
  number={1},
  pages={7229},
  year={2022},
  publisher={Nature Publishing Group UK London}
}

@article{team2025gemma,
  title={Gemma 3 technical report},
  author={Team, Gemma and Kamath, Aishwarya and Ferret, Johan and Pathak, Shreya and Vieillard, Nino and Merhej, Ramona and Perrin, Sarah and Matejovicova, Tatiana and Ram{\'e}, Alexandre and Rivi{\`e}re, Morgane and others},
  journal={arXiv preprint arXiv:2503.19786},
  year={2025}
}

@inproceedings{llm4hypergraph,
  title={Beyond Graphs: Can Large Language Models Comprehend Hypergraphs?},
  author={Feng, Yifan and Yang, Chengwu and Hou, Xingliang and Du, Shaoyi and Ying, Shihui and Wu, Zongze and Gao, Yue},
  booktitle={The Thirteenth International Conference on Learning Representations},
year = {2025}
}

@article{li2025benchmark,
  title={Benchmark evaluations, applications, and challenges of large vision language models: A survey},
  author={Li, Zongxia and Wu, Xiyang and Du, Hongyang and Nghiem, Huy and Shi, Guangyao},
  journal={arXiv preprint arXiv:2501.02189},
  volume={1},
  year={2025}
}

@inproceedings{yang2016revisiting,
  title={Revisiting semi-supervised learning with graph embeddings},
  author={Yang, Zhilin and Cohen, William and Salakhudinov, Ruslan},
  booktitle={International conference on machine learning},
  pages={40--48},
  year={2016},
  organization={PMLR}
}

@article{yadati2019hypergcn,
  title={Hypergcn: A new method for training graph convolutional networks on hypergraphs},
  author={Yadati, Naganand and Nimishakavi, Madhav and Yadav, Prateek and Nitin, Vikram and Louis, Anand and Talukdar, Partha},
  journal={Advances in neural information processing systems},
  volume={32},
  year={2019}
}

@article{m-csa,
  title={Mechanism and Catalytic Site Atlas (M-CSA): a database of enzyme reaction mechanisms and active sites},
  author={Ribeiro, Ant{\'o}nio J M and Holliday, Gemma L and Furnham, Nicholas and Tyzack, Jonathan D and Ferris, Katherine and Thornton, Janet M},
  journal={Nucleic acids research},
  volume={46},
  number={D1},
  pages={D618--D623},
  year={2018},
  publisher={Oxford University Press}
}

@inproceedings{pubmed,
  title={Revisiting semi-supervised learning with graph embeddings},
  author={Yang, Zhilin and Cohen, William and Salakhudinov, Ruslan},
  booktitle={International conference on machine learning},
  pages={40--48},
  year={2016},
  organization={PMLR}
}

@inproceedings{vgcure,
  title={Benchmarking and improving large vision-language models for fundamental visual graph understanding and reasoning},
  author={Zhu, Yingjie and Bai, Xuefeng and Chen, Kehai and Xiang, Yang and Yu, Jun and Zhang, Min},
  booktitle={Proceedings of the 63rd Annual Meeting of the Association for Computational Linguistics (Volume 1: Long Papers)},
  pages={30678--30701},
  year={2025}
}

@article{ravetz2019photoredox,
  title={Photoredox catalysis using infrared light via triplet fusion upconversion},
  author={Ravetz, Benjamin D and Pun, Andrew B and Churchill, Emily M and Congreve, Daniel N and Rovis, Tomislav and Campos, Luis M},
  journal={Nature},
  volume={565},
  number={7739},
  pages={343--346},
  year={2019},
  publisher={Nature Publishing Group UK London}
}

@article{feng2025hyperrag,
  title={Hyper-RAG: Combating LLM Hallucinations using Hypergraph-Driven Retrieval-Augmented Generation},
  author={Feng, Yifan and Hu, Hao and Hou, Xingliang and Liu, Shiquan and Ying, Shihui and Du, Shaoyi and Hu, Han and Gao, Yue},
  journal={arXiv preprint arXiv:2504.08758},
  year={2025}
}

@article{dijkstra2022note,
  title={A note on two problems in connexion with graphs},
  author={Dijkstra, Edsger W},
  journal={Numerische Mathematik},
  year={1959}
}

@article{wei2024gita,
  title={Gita: Graph to visual and textual integration for vision-language graph reasoning},
  author={Wei, Yanbin and Fu, Shuai and Jiang, Weisen and Zhang, Zejian and Zeng, Zhixiong and Wu, Qi and Kwok, James and Zhang, Yu},
  journal={Advances in Neural Information Processing Systems},
  volume={37},
  pages={44--72},
  year={2024}
}

@article{li2024visiongraph,
  title={Visiongraph: Leveraging large multimodal models for graph theory problems in visual context},
  author={Li, Yunxin and Hu, Baotian and Shi, Haoyuan and Wang, Wei and Wang, Longyue and Zhang, Min},
  journal={arXiv preprint arXiv:2405.04950},
  year={2024}
}

@techreport{gpt4o,
    author = {OpenAI},
    title = {{GPT-4o}},
    type = {Technical Report},
    institution  = {OpenAI},
    year = {2024}
}

@techreport{gemini3,
    author = {Google},
    title = {{Gemini-3 Pro}},
    type = {Technical Report},
    institution = {Google DeepMind},
    year = {2025}
}

@inproceedings{deitke2025molmo,
  title={Molmo and pixmo: Open weights and open data for state-of-the-art vision-language models},
  author={Deitke, Matt and Clark, Christopher and Lee, Sangho and Tripathi, Rohun and Yang, Yue and Park, Jae Sung and Salehi, Mohammadreza and Muennighoff, Niklas and Lo, Kyle and Soldaini, Luca and others},
  booktitle={Proceedings of the Computer Vision and Pattern Recognition Conference},
  pages={91--104},
  year={2025}
}

@misc{liu2024llavanext,
    title={LLaVA-NeXT: Improved reasoning, OCR, and world knowledge},
    url={https://llava-vl.github.io/blog/2024-01-30-llava-next/},
    author={Liu, Haotian and Li, Chunyuan and Li, Yuheng and Li, Bo and Zhang, Yuanhan and Shen, Sheng and Lee, Yong Jae},
    month={January},
    year={2024}
}

@misc{llava1.5,
      title={Improved Baselines with Visual Instruction Tuning}, 
      author={Liu, Haotian and Li, Chunyuan and Li, Yuheng and Lee, Yong Jae},
      publisher={arXiv:2310.03744},
      year={2023},
}

@book{matplotlib,
  title={Matplotlib for Python developers},
  author={Tosi, Sandro},
  year={2009},
  publisher={Packt Publishing Ltd}
}

@techreport{networkx,
  title={Exploring network structure, dynamics, and function using NetworkX},
  author={Hagberg, Aric and Swart, Pieter and S Chult, Daniel},
  year={2008},
  institution={Los Alamos National Lab.(LANL), Los Alamos, NM (United States)}
}

@book{berge1985graphs,
  title={Graphs and hypergraphs},
  author={Berge, Claude},
  year={1985},
  publisher={Elsevier Science Ltd.}
}

@book{voloshin2002coloring,
  title={Coloring Mixed Hypergraphs: Theory, Algorithms and Applications: Theory, Algorithms, and Applications},
  author={Voloshin, Vitaly Ivanovich},
  number={17},
  year={2002},
  publisher={American Mathematical Soc.}
}

@inproceedings{babai2016graph,
  title={Graph isomorphism in quasipolynomial time},
  author={Babai, L{\'a}szl{\'o}},
  booktitle={Proceedings of the forty-eighth annual ACM symposium on Theory of Computing},
  pages={684--697},
  year={2016}
}

@article{garey1990guide,
  title={A Guide to the Theory of NP-Completeness},
  author={Garey, Michael R and Johnson, David S and others},
  journal={Computers and intractability},
  pages={37--79},
  year={1990},
  publisher={Freeman}
}

@article{bretto2013hypergraph,
  title={Hypergraph theory},
  author={Bretto, Alain},
  journal={An introduction. Mathematical Engineering. Cham: Springer},
  volume={1},
  pages={209--216},
  year={2013},
  publisher={Springer}
}

@article{scipy,
  title={SciPy 1.0: fundamental algorithms for scientific computing in Python},
  author={Virtanen, Pauli and Gommers, Ralf and Oliphant, Travis E and Haberland, Matt and Reddy, Tyler and Cournapeau, David and Burovski, Evgeni and Peterson, Pearu and Weckesser, Warren and Bright, Jonathan and others},
  journal={Nature methods},
  volume={17},
  number={3},
  pages={261--272},
  year={2020},
  publisher={Nature Publishing Group US New York}
}

@incollection{bermond1978hamiltonian,
  title={Hamiltonian decompositions of graphs, directed graphs and hypergraphs},
  author={Bermond, J-C},
  booktitle={Annals of Discrete Mathematics},
  volume={3},
  pages={21--28},
  year={1978},
  publisher={Elsevier}
}

@article{dacar1998cyclicity,
  title={The cyclicity of a hypergraph},
  author={Dacar, France},
  journal={Discrete Mathematics},
  volume={182},
  number={1-3},
  pages={53--67},
  year={1998},
  publisher={Elsevier}
}

@book{ahuja1994network,
  title={Network flows: theory, algorithms and applications},
  author={Ahuja, Ravindra K and Magnanti, Thomas L and Orlin, James B},
  year={1994},
  publisher={Prentice hall}
}

@article{edmonds1972theoretical,
  title={Theoretical improvements in algorithmic efficiency for network flow problems},
  author={Edmonds, Jack and Karp, Richard M},
  journal={Journal of the ACM (JACM)},
  volume={19},
  number={2},
  pages={248--264},
  year={1972},
  publisher={ACM New York, NY, USA}
}

@article{he2021debertav3,
  title={Debertav3: Improving deberta using electra-style pre-training with gradient-disentangled embedding sharing},
  author={He, Pengcheng and Gao, Jianfeng and Chen, Weizhu},
  journal={arXiv preprint arXiv:2111.09543},
  year={2021}
}

@article{internvl2.5,
    title={Expanding Performance Boundaries of Open-Source Multimodal Models with Model, Data, and Test-Time Scaling},
    author={Chen, Zhe and Wang, Weiyun and Cao, Yue and Liu, Yangzhou and Gao, Zhangwei and Cui, Erfei and Zhu, Jinguo and Ye, Shenglong and Tian, Hao and Liu, Zhaoyang and others},
    journal={arXiv preprint arXiv:2412.05271},
    year={2024}
}

@article{zhu2025internvl3,
  title={Internvl3: Exploring advanced training and test-time recipes for open-source multimodal models},
  author={Zhu, Jinguo and Wang, Weiyun and Chen, Zhe and Liu, Zhaoyang and Ye, Shenglong and Gu, Lixin and Tian, Hao and Duan, Yuchen and Su, Weijie and Shao, Jie and others},
  journal={arXiv preprint arXiv:2504.10479},
  year={2025}
}

@article{bylander1994computational,
  title={The computational complexity of propositional STRIPS planning},
  author={Bylander, Tom},
  journal={Artificial Intelligence},
  volume={69},
  number={1-2},
  pages={165--204},
  year={1994},
  publisher={Elsevier}
}

@article{glm2024chatglm,
  title={Chatglm: A family of large language models from glm-130b to glm-4 all tools},
  author={GLM, Team and Zeng, Aohan and Xu, Bin and Wang, Bowen and Zhang, Chenhui and Yin, Da and Zhang, Dan and Rojas, Diego and Feng, Guanyu and Zhao, Hanlin and others},
  journal={arXiv preprint arXiv:2406.12793},
  year={2024}
}

@inproceedings{ai2024advancement,
  title={Advancement in graph understanding: A multimodal benchmark and fine-tuning of vision-language models},
  author={Ai, Qihang and Li, Jiafan and Dai, Jincheng and Zhou, Jianwu and Liu, Lemao and Jiang, Haiyun and Shi, Shuming},
  booktitle={Proceedings of the 62nd Annual Meeting of the Association for Computational Linguistics (Volume 1: Long Papers)},
  pages={7485--7501},
  year={2024}
}

@article{ruggeri2023community,
  title={Community detection in large hypergraphs},
  author={Ruggeri, Nicol{\`o} and Contisciani, Martina and Battiston, Federico and De Bacco, Caterina},
  journal={Science Advances},
  volume={9},
  number={28},
  pages={eadg9159},
  year={2023},
  publisher={American Association for the Advancement of Science}
}

@article{Qwen3-VL,
      title={Qwen3-VL Technical Report}, 
      author={Shuai Bai and Yuxuan Cai and Ruizhe Chen and Keqin Chen and Xionghui Chen and Zesen Cheng and Lianghao Deng and Wei Ding and Chang Gao and Chunjiang Ge and Wenbin Ge and Zhifang Guo and Qidong Huang and Jie Huang and Fei Huang and Binyuan Hui and Shutong Jiang and Zhaohai Li and Mingsheng Li and Mei Li and Kaixin Li and Zicheng Lin and Junyang Lin and Xuejing Liu and Jiawei Liu and Chenglong Liu and Yang Liu and Dayiheng Liu and Shixuan Liu and Dunjie Lu and Ruilin Luo and Chenxu Lv and Rui Men and Lingchen Meng and Xuancheng Ren and Xingzhang Ren and Sibo Song and Yuchong Sun and Jun Tang and Jianhong Tu and Jianqiang Wan and Peng Wang and Pengfei Wang and Qiuyue Wang and Yuxuan Wang and Tianbao Xie and Yiheng Xu and Haiyang Xu and Jin Xu and Zhibo Yang and Mingkun Yang and Jianxin Yang and An Yang and Bowen Yu and Fei Zhang and Hang Zhang and Xi Zhang and Bo Zheng and Humen Zhong and Jingren Zhou and Fan Zhou and Jing Zhou and Yuanzhi Zhu and Ke Zhu},
	  journal={arXiv preprint arXiv:2511.21631},
      year={2025}
}

@article{deepseekvl2,
  title={Deepseek-vl2: Mixture-of-experts vision-language models for advanced multimodal understanding},
  author={Wu, Zhiyu and Chen, Xiaokang and Pan, Zizheng and Liu, Xingchao and Liu, Wen and Dai, Damai and Gao, Huazuo and Ma, Yiyang and Wu, Chengyue and Wang, Bingxuan and others},
  journal={arXiv preprint arXiv:2412.10302},
  year={2024}
}

@article{qwen25vl,
  title={Qwen2. 5-vl technical report},
  author={Bai, Shuai and Chen, Keqin and Liu, Xuejing and Wang, Jialin and Ge, Wenbin and Song, Sibo and Dang, Kai and Wang, Peng and Wang, Shijie and Tang, Jun and others},
  journal={arXiv preprint arXiv:2502.13923},
  year={2025}
}

@inproceedings{li2025heal,
  title={HEAL: Hybrid Enhancement with LLM-based Agents for Text-attributed Hypergraph Self-supervised Representation Learning},
  author={Li, Ruochang and Luo, Xiao and Xiao, Zhiping and Ju, Wei and Zhang, Ming},
  booktitle={Findings of the Association for Computational Linguistics: EMNLP 2025},
  pages={6815--6829},
  year={2025}
}

@article{chu2024llm,
  title={Llm-guided multi-view hypergraph learning for human-centric explainable recommendation},
  author={Chu, Zhixuan and Wang, Yan and Cui, Qing and Li, Longfei and Chen, Wenqing and Qin, Zhan and Ren, Kui},
  journal={arXiv preprint arXiv:2401.08217},
  year={2024}
}


\clearpage

\appendix

\section{Related Works}
\label{app:related}
\paragraph{Benchmarking LVLMs on Graphs.}
VisionGraph \citep{li2024visiongraph} and GVLQA \citep{wei2024gita} evaluate the problem-solving capabilities of LVLMs in graph theory problems. VisionGraph assumes that visual graphs are naturally equipped, while GVLQA generates visual graphs from scratch. Both benchmarks include numerous synthetic visual graphs and complex graph theory problems. \citet{ai2024advancement} introduces a multimodal instruction-following benchmark designed to assess LVLMs' understanding and reasoning capabilities on real-world graph images across various domains. \textsc{VGCure} \citep{vgcure} establishes fundamental understanding and reasoning benchmarks tailored to explore how LVLMs understand and reason on visual graphs. 

However, these benchmarks are limited to ordinary graphs, leaving LVLMs' capabilities in handling high-order relationships in hypergraphs unexplored. Hypergraphs have substantial practical value in domains such as retrieved-augmented generation (RAG) \citep{feng2025hyperrag}, life sciences \citep{ravetz2019photoredox}, and community analysis \citep{ruggeri2023community}. To address this gap, we propose \texttt{HyperGVL}, the first dedicated benchmark for evaluating LVLMs' capabilities in hypergraph understanding and reasoning.

\paragraph{Large Models for Hypergraphs}

In the realm of benchmarking, the sole existing benchmark for evaluating large models' hypergraph capabilities is \texttt{LLM4Hypergraph}, which rigorously assesses LLMs' proficiency in understanding hypergraph structures. Our proposed \texttt{HyperGVL} diverges from \texttt{LLM4Hypergraph} in two pivotal aspects: 1) \texttt{HyperGVL} is tailored for LVLMs, where visual components complement textual information, yielding unique informational gains and expanding capability boundaries. 2) We have augmented task complexity and diversity, demanding models to engage in sophisticated reasoning based on hypergraph comprehension. This approach not only meets the evolving requisites of advanced models but also ensures comprehensive coverage of difficulty levels and evaluation capabilities.

Beyond benchmark creation, recent studies have delved into the role of large models in hypergraph-related learning and reasoning. HEAL~\cite{li2025heal} explores the use of LLM-based agents to enhance representation learning on text-attributed hypergraphs, positioning large language models as tools to improve data quality and structural signals under limited supervision. LLaSA~\cite{llasa} designs a hypergraph-aware LLM for knowledge grounding, integrating unified hypergraph representations into LLMs. LLMHG~\cite{chu2024llm} employs large language models to generate structured user interest angles, facilitating multi-view hypergraph construction. These advancements collectively underscore the transformative potential of large models in hypergraph processing and their growing importance in the field.

\section{Hypergraph Generation}
In this section, we introduce the hypergraph generation in our \texttt{HyperGVL} with details.
\label{sec:structure}
\subsection{Synthetic Hypergraph}
\label{app:synthetic_hypergraphs}

Synthetic hypergraphs are generated using our custom-developed scripts. For most tasks, we employ a randomized construction approach that creates connected hypergraphs by iteratively adding hyperedges of varying sizes and ensuring connectivity through breadth-first traversal. For NP-hard reasoning tasks such as SHC, HHM, and 3-CL, we utilize specialized constructors designed to ensure the existence of valid solutions. SHC instances are constructed with explicit hypercycle structures where consecutive hyperedges share exactly one vertex. HHM instances are formed using vertex permutations as backbone paths. Meanwhile, 3-CL instances are generated from pre-assigned color classes with constraint-preserving hyperedge selection.
These scripts will be released together with our benchmark.

\subsection{Real-world Hypergraph}
\label{app:real_hypergraphs}

Real-world hypergraphs are derived from the citation co-author network PubMed\_CA \cite{pubmed} and protein structure resources M-CSA \cite{m-csa}. In these hypergraphs, vertices represent anonymized authors or residues, while hyperedges correspond to paper-level co-author relationships or catalytic-site residue sets, respectively. To create scale-controlled instances while preserving real structural patterns, we extract sub-hypergraphs from the full hypergraphs by selecting a seed vertex and traversing incident hyperedges using a random walk, retaining visited vertices until reaching the target size. Finally, all vertex and hyperedge identifiers are re-indexed to canonical IDs. For tasks such as 3-CL, SHC, and HHM, which require the hypergraph structure to ensure a valid solution exists, we repeat the sampling until the sampled hypergraph meets the requirements.

\subsection{Hypergraph Structure Control}

We employ the scale partition protocol from \citet{llm4hypergraph}, categorizing hypergraphs by vertex count into three groups: \textit{small} (5–10 vertices), \textit{medium} (10–15 vertices), and \textit{large} (15–20 vertices). The dataset consists of synthetic and real-world hypergraphs in a 1:1 ratio, with an internal distribution of small:medium:large set at 1:2:1. This classification aids in evaluating performance across increasingly complex scenarios.

For synthetic hypergraphs, hyperedge density is regulated by ensuring the number of hyperedges falls within the range of $[0.2|V|, 1.5|V|]$. This constraint prevents the generation of structures that are either trivially sparse or overly dense, enabling more informative capability assessments. Conversely, real-world hypergraphs do not impose a hyperedge-count constraint, allowing their hyperedge statistics to naturally reflect the connectivity patterns inherent in the original data.

\begin{table*}[t]
	\caption{  \label{tab:realset}
		Dataset Statistics for Zero-shot Hypergraph Node Classification.}
	\centering
	\small

	\begin{tabular}{l|l|l|l|l|l}
		\hline
		& \textbf{DBLP-CA} & \textbf{Pubmed-CC} & \textbf{Cora-CA} &\textbf{Cora-CC} & \textbf{Citeseer-CC}  \\
		& (co-authorship) & (co-citation)  & (co-authorship) & (co-citation) & (co-citation) \\
		\hline
		\# nodes(vertices), $|V|$ & $43413$ & $19717$ & $2708$   & $2708$ & $3312$ \\
		\# hyperedges, $|E|$ & $22535$  & $7963$ & $1072$ & $1579$ & $1079$ \\
		avg.hyperedge order & $4.7\pm6.1$ & $4.3\pm5.7$ & $4.2\pm4.1$ & $3.0\pm1.1$ & $3.2\pm2.0$ \\		
        \# node classes
        & 6 & 3 & 7 & 7 & 6\\
		\hline
	\end{tabular}
\end{table*}

\section{Task Details}
\label{app:task}
In this section, we detail the 12 tasks in \texttt{HyperGVL}, categorized by their primary focus (understanding vs. reasoning) and difficulty level. Each task includes a formal problem definition, evaluation methodology, and implementation details from our codebase.

\subsection{Understanding Tasks}
Understanding tasks assess three core abilities of hypergraph comprehension for LVLMs: (1) \textit{basic element capture}, recognizing vertices and hyperedges, (2) \textit{adjacency perception}, understanding vertex adjacency relationships, and (3) \textit{heuristic computation}, calculating heuristics like vertex degrees and hyperedge orders.

\ 

\noindent{\bf Level-1 tasks} focus on individual basic element capture or adjacency perception, serving as foundational components for other understanding and reasoning tasks.

\begin{itemize}
    \item \textbf{Vertex Counting (VC).} This task tests the model's ability to count vertices in a hypergraph. Given a hypergraph $\mathcal{G} = \{V, E\}$, the goal is to return $|V|$, the vertex set's cardinality. Evaluation is based on exact matches with pre-computed ground truth.

    \item \textbf{Hyperedge Counting (HEC).} This task assesses the model's ability to count hyperedges in a hypergraph. Given a hypergraph $\mathcal{G} = \{V, E\}$, the goal is to return $|E|$, the hyperedge set's cardinality. Unlike ordinary graph edge counting, this requires distinguishing each hyperedge from multiple connected vertices. Evaluation is based on exact matches with pre-computed ground truth.

    \item \textbf{Neighbors (Ne).} This task evaluates the model's ability to identify all direct neighbors of a given vertex. Given $\mathcal{G} = \{V, E\}$ and a vertex $u \in V$, the goal is to output the subset $\mathcal{N}(u) \subseteq V$ such that for each $v \in \mathcal{N}(u)$, there exists a hyperedge $e \in E$ where $u \in e$ and $v \in e$. Evaluation is performed by set-based comparison, allowing any ordering between the LVLMs' response and the pre-computed ground truth set.
\end{itemize}

\noindent{\bf Level-2 tasks} integrate heuristic computations, such as vertex degrees and hyperedge orders, into Level-1 tasks. This combination increases task complexity by simultaneously requiring the accomplishment of compound atomic capabilities.

\begin{itemize}
    \item \textbf{Degree-specified Vertex Counting (DVC).} This task requires counting vertices that meet a specific degree constraint. Given a hypergraph $\mathcal{G} = \{V, E\}$ and a target degree $d$, the goal is to return the number of vertices $v \in V$ such that $\text{deg}(v) = |\{e \in E \mid v \in e\}| = d$. The target degree is randomly selected from the hypergraph's existing degree distribution, ensuring task validity. Evaluation is based on exact matches with pre-computed ground truth.

    \item \textbf{Order-specified Edge Counting (OEC).} This task evaluates the model's understanding of hyperedge order (cardinality). Given a hypergraph $\mathcal{G} = \{V, E\}$ and a target order $k$, the goal is to count hyperedges $e \in E$ where $|e| = k$. The hyperedge order represents the number of vertices it contains, a unique characteristic of hypergraphs. The target order is randomly sampled from the existing order distribution in the hypergraph. Evaluation is based on exact matches with pre-computed ground truth.

    \item \textbf{Order-filtered Neighbors (ONe).} This task extends basic neighbor identification by introducing order-based filtering. Given a hypergraph $\mathcal{G} = \{V, E\}$, a vertex $u$, and a minimum order threshold $k$, the goal is to identify neighbors of $u$ connected only through hyperedges with order $\ge k$. Formally, return $\{v \in V \mid \exists e \in E : u \in e \land v \in e \land |e| \ge k\}$. Evaluation is performed by set-based comparison, allowing any ordering between the LVLMs' response and the pre-computed ground truth set.
\end{itemize}

\subsection{Reasoning Tasks}
Reasoning tasks represent an advanced stage beyond understanding tasks, requiring coherent, multi-step reasoning to analyze and manipulate the information acquired from structural understanding.

\ 

\noindent{\bf Level-3 tasks} involve reasoning on problems with known polynomial-time or quasi-polynomial-time algorithms. Due to the necessity of following strict algorithmic steps, where each step examines individual or compound comprehension abilities and requires organization according to the algorithm's sequence, these tasks present a greater challenge.

\begin{itemize}
    \item \textbf{Order-weighted Shortest Path (OSP).} This task evaluates the model's ability to compute shortest paths in weighted hypergraphs. Given a hypergraph $\mathcal{G} = \{V, E\}$ and vertices $s, t \in V$, where each hyperedge $e$ has weight $w(e) = |e|$ (its order), the goal is to find the minimum total weight $W$ of a path from $s$ to $t$. A path is a sequence of hyperedges where consecutive hyperedges share at least one vertex. Implementation uses a Dijkstra-like algorithm adapted for hypergraphs \cite{dijkstra2022note}. Evaluation is based on exact matches with pre-computed ground truth.

    \item \textbf{Order-weighted Maximum Flow (OMF).} This task assesses the model's capability to estimate maximum flow in hypergraph networks. Given a hypergraph $\mathcal{G} = \{V, E\}$ with source $s$ and sink $t$, where each hyperedge $e$ has capacity $c(e) = |e|$, the goal is to compute the maximum flow from $s$ to $t$. The hypergraph is converted to a flow network by constructing its bipartite incidence graph \cite{berge1985graphs}: we introduce a dedicated intermediate node for each hyperedge, and connect each intermediate node to all vertices contained in the corresponding hyperedge. To adapt to maximum flow computation \cite{ahuja1994network}, edges between intermediate nodes and their associated vertices are assigned capacities equal to the hyperedge's capacity \( c(e) = |e| \). The Edmonds-Karp algorithm \cite{edmonds1972theoretical} is then applied to this transformed flow network. Evaluation is based on exact matches with pre-computed ground truth.

    \item \textbf{Isomorphism Recognition (ISM).} This task evaluates the model's ability to recognize structural equivalence between hypergraphs. Given two hypergraphs $G = \{V, E\}$ and $H = \{V', E'\}$, the goal is to determine whether a bijective mapping $f: V \rightarrow V'$ exists such that $(v_1, \ldots, v_k) \in E \iff (f(v_1), \ldots, f(v_k)) \in E'$ for all hyperedges \cite{babai2016graph}. The benchmark generates both positive (isomorphic via vertex relabeling) and negative (structurally different) examples with equal probability. Evaluation is based on binary classification accuracy.
\end{itemize}

\noindent{\textbf{Level-4 Tasks}} require reasoning on NP-hard (NP-complete) \cite{garey1990guide} problems, which are considered classic and extremely challenging. These tasks lack definitive polynomial-time algorithms as guides, often necessitating demanding models that proactively search and independently plan for feasible solutions.

\begin{itemize}
    \item \textbf{Hypergraph 3-Coloring (3-CL).} This task assesses the model's planning capability on the NP-complete hypergraph coloring problem \cite{voloshin2002coloring}, a generalization of the classic graph 3-coloring problem to high-order hypergraph structures. Given a hypergraph $\mathcal{G} = \{V, E\}$, determine if there exists a coloring $c: V \rightarrow \{c_0, c_1, c_2\}$ such that every hyperedge $e \in E$ satisfies $|\{c(v) \mid v \in e\}| \ge  2$ (contains at least 2 different colors). Each case includes at least one feasible coloring strategy. Solutions are verified using automated scripts.

    \item \textbf{Strict Hypercycle (SHC).} This task evaluates cycle detection and validation in hypergraphs with strict intersection constraints. A strict hypercycle is a sequence of hyperedges $e_1, e_2, \ldots, e_k$ ($k \ge 2$) where $|e_i \cap e_{i+1}| = 1$ for $i = 1, \ldots, k-1$ and $|e_k \cap e_1| = 1$ (consecutive hyperedges share exactly one vertex, forming a closed loop) \cite{bretto2013hypergraph,dacar1998cyclicity}. Given $\mathcal{G} = \{V, E\}$, determine if such a cycle exists and provide it. Solutions are verified using automated scripts.

    \item \textbf{Hypergraph Hamiltonian Path (HHM).} This task tests the model's ability to solve the NP-complete Hamiltonian path problem on hypergraphs \cite{bermond1978hamiltonian}, an extension of the classic graph Hamiltonian path problem to hypergraphs. Given $\mathcal{G} = \{V, E\}$ and vertices $s, t \in V$, determine if there exists a sequence of hyperedges $e_1, \ldots, e_m$ that forms a path from $s$ to $t$ visiting every vertex in $V$ exactly once. The path is represented as edge indices rather than vertex sequences. Solutions are verified using automated scripts.
\end{itemize}

\begin{figure*}[t]
    \centering
    \includegraphics[width=\linewidth]{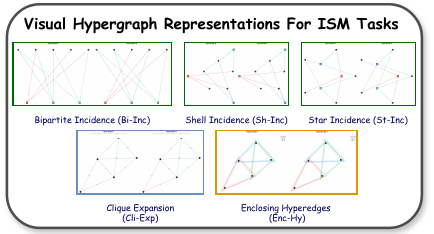}
    \caption{The visual hypergraph representations for the ISM task. The similar layouts between the two hypergraph visualizations provide straightforward cues for distinguishing isomorphism.}
    \label{fig:ism}
\end{figure*}

\section{Visual Representation Generation}
\label{app:visual}
Visual hypergraph representations in \texttt{HyperGVL} are generated using custom visualization algorithms implemented with NetworkX \cite{networkx} and Matplotlib \cite{matplotlib}. Unlike traditional graph visualization methods, hypergraph visualization must explicitly represent hyperedges that connect arbitrary numbers of vertices, requiring specialized encoding strategies. We implement five distinct visual representation families, each designed to highlight different structural aspects of hypergraphs. The examples of each type have been provided in Fig.~\ref{fig:language}.

\subsection{Implementation Details}

All visualizations transform hypergraph structures into images with resolution 1400$\times$1100 pixels at 150 DPI. The implementation uses the following core transformations:

\paragraph{Bipartite Transformation.} For V-E pairwise representations (Bi-Inc, Sh-Inc, St-Inc), hypergraphs are first converted to bipartite graphs $G_{\text{bi}} = (V \cup E, E_{\text{bi}})$, where vertices and hyperedges form two disjoint node sets, and edges $E_{\text{bi}}$ represent membership relations. Formally, for each hyperedge $e_i \in E$ and vertex $v_j \in e_i$, we create an edge $(v_j, e_i) \in E_{\text{bi}}$. This bipartite structure is then visualized using different layout algorithms:

\begin{itemize}
\item \textbf{Bipartite Incidence (Bi-Inc)}: Employs a horizontal two-layer layout where vertices are positioned in the top row and hyperedge nodes in the bottom row, with straight lines depicting membership. Vertex nodes are rendered as black circles ($\circ$, radius=20pt) with white labels, while hyperedge nodes are colored squares ($\Box$, size=18pt) with distinct colors assigned to each hyperedge.

\item \textbf{Shell Incidence (Sh-Inc)}: Uses NetworkX's \texttt{shell\_layout} algorithm to arrange vertices in an inner concentric circle and hyperedge nodes in an outer circle. This radial arrangement facilitates the identification of vertex-hyperedge connectivity patterns through visual clustering.

\item \textbf{Star Incidence (St-Inc)}: Similar to Sh-Inc but with vertices positioned on an outer circle and hyperedge nodes placed at the center, with radial connections emanating from hyperedge centers to member vertices. This star-like pattern emphasizes the grouping structure of each hyperedge.
\end{itemize}

\paragraph{Clique Expansion.} For Cli-Exp representation, each hyperedge $e = \{v_1, v_2, \ldots, v_k\}$ is decomposed into a complete subgraph over its vertices, generating $\binom{k}{2}$ pairwise edges. Edge labels annotate each pairwise connection with the hyperedge IDs it belongs to (e.g., ``e0,e2'' indicates the edge participates in both hyperedge 0 and 2). Layout is computed using NetworkX's \texttt{spring\_layout} with parameters $k=2.0$, iterations=100, scale=3.0 to minimize edge crossings while maintaining readability.

\paragraph{Enclosing Regions.} For Enc-Hy representation, we implement a labeled convex hull visualization using NetworkX's \texttt{kamada\_kawai\_layout} for vertex positioning, which minimizes graph energy to reduce overlaps. For each hyperedge $e_i$ with $|e_i| \geq 3$, we compute the convex hull of its member vertices using SciPy's \texttt{ConvexHull} \cite{scipy}, rendering it as a colored polygonal region with 15\% opacity. Crucially, a white rectangular label box displaying the hyperedge ID (e.g., ``e2'') is placed at the centroid of the region to provide explicit identification, addressing the perceptual ambiguity of color-only encoding.

\subsection{Design Rationale}

The choice of these five representations is motivated by three design principles validated through preliminary LVLM experiments: (1) \textit{Explicitness}: Hyperedges must have visual identifiers beyond color coding, as LVLMs struggle with pure color-based semantic extraction. All representations except Enc-Hy use explicit hyperedge nodes or edge labels. (2) \textit{Structural diversity}: Different representations emphasize complementary structural properties: Bipartite views highlight membership relations, clique expansion reveals pairwise connectivity, and convex hulls show geometric groupings. (3) \textit{Scalability}: Layouts are optimized for small-to-medium hypergraphs (5–20 vertices) to maintain visual clarity and avoid information overload.

\subsection{Visual Representations for ISM task}
\label{app:ism}

Isomorphism Recognition (ISM) is special among the tasks, as it contains two hypergraphs in one query. 
For this task, we observe that isomorphic hypergraphs often exhibit similar visual patterns under deterministic layouts (with fixed random seeds), as layout algorithms preserve structural symmetries. This phenomenon contributes to the higher performance on ISM than our expectation, as illustrated in Tab.\ref{tab:QA_results}. Figure \ref{fig:ism} illustrates various visual hypergraph representations used in ISM tasks. In these representations, most layouts can directly reveal isomorphism relationships through their striking similarities.

\section{Prompt Examples}

In this section, we exemplified the prompts used in \texttt{HyperGVL} evaluation.

\subsection{Textual Representation Prompts}
\label{app:prompt_t}
We provide representative examples of textual hypergraph representations used in \texttt{HyperGVL}. The following prompts demonstrate different encoding approaches for a simple hypergraph with 5 vertices (v0--v4) and 3 hyperedges (e0--e2), where e0=\{v0, v1, v2\}, e1=\{v1, v2, v3\}, and e2=\{v2, v3, v4\}.


\begin{myprompt}[label=prompt:lo-inc]{LO-Inc}
    \textbf{Prompt: }\textit{G describes a hypergraph among vertices v0, v1, v2, v3, and v4 and hyperedges e0, e1, and e2.\\
    In this hypergraph:\\
    Vertex v0 is connected to vertices v1, v2.\\
    Vertex v1 is connected to vertices v0, v2, v3.\\
    Vertex v2 is connected to vertices v0, v1, v3, v4.\\
    Vertex v3 is connected to vertices v1, v2, v4.\\
    Vertex v4 is connected to vertices v2, v3.}
\end{myprompt}

\begin{myprompt}[label=prompt:n-pair]{N-Pair}
    \textbf{Prompt: }\textit{In an undirected hypergraph, (i,j) means that vertex i and vertex j are connected with an undirected hyperedge. G describes a hypergraph among vertices v0, v1, v2, v3, and v4 and hyperedges e0, e1, and e2.\\
    The connection relation between vertices in G are: (v0, v1) (v0, v2) (v1, v2) (v1, v3) (v2, v3) (v2, v4) (v3, v4).}
\end{myprompt}

\begin{myprompt}[label=prompt:adj-mat]{Adj-Mat}
    \textbf{Prompt: }\textit{G describes a hypergraph among vertices v0, v1, v2, v3, and v4 and among hyperedges e0, e1, and e2.\\
    The adjacency matrix between the vertices of the hypergraph is\newline
    [[0,1,1,0,0,]\newline, 
    [1,0,1,1,0,]\newline, 
    [1,1,0,1,1,]\newline, 
    [0,1,1,0,1,]\newline, 
    [0,0,1,1,0,]]}
\end{myprompt}


\begin{myprompt}[label=prompt:ho-neigh]{HO-Neigh}
    \textbf{Prompt: }\textit{G describes a hypergraph among vertices v0, v1, v2, v3, and v4 and hyperedges e0, e1, and e2.\\
    In this hypergraph:\\
    Vertex v0 is connected to hyperedge e0.\\
    Vertex v1 is connected to hyperedges e0, e1.\\
    Vertex v2 is connected to hyperedges e0, e1, e2.\\
    Vertex v3 is connected to hyperedges e1, e2.\\
    Vertex v4 is connected to hyperedge e2.\\
    Hyperedge e0 is connected to vertices v0, v1, v2.\\
    Hyperedge e1 is connected to vertices v1, v2, v3.\\
    Hyperedge e2 is connected to vertices v2, v3, v4.}
\end{myprompt}

\begin{myprompt}[label=prompt:inc-mat]{Inc-Mat}
    \textbf{Prompt: }\textit{G describes a hypergraph among vertices v0, v1, v2, v3, and v4 and hyperedges e0, e1, and e2.\\
    The incidence matrix of the hypergraph is\newline
    [[1,0,0,],\newline
    [1,1,0,],\newline
    [1,1,1,],\newline
    [0,1,1,],\newline
    [0,0,1,]]}
\end{myprompt}


\begin{myprompt}[label=prompt:n-set]{N-Set}
    \textbf{Prompt: }\textit{In an undirected hypergraph, (i, j, k) means that vertex i, vertex j, and vertex k are connected with an undirected hyperedge. G describes a hypergraph among vertices v0, v1, v2, v3, and v4, and among hyperedges e0, e1, and e2.\\
    The hyperedges in G are: (v0, v1, v2), (v1, v2, v3), (v2, v3, v4).}
\end{myprompt}

\begin{myprompt}[label=prompt:ho-inc]{HO-Inc}
    \textbf{Prompt: }\textit{G describes a hypergraph among vertices v0, v1, v2, v3, and v4 and among hyperedges e0, e1, and e2.\\
    In this hypergraph:\\
    Vertex v0 is connected to vertices v1, v2 with hyperedge e0.\\
    Vertex v1 is connected to vertices v0, v2 with hyperedge e0, to vertices v2, v3 with hyperedge e1.\\
    Vertex v2 is connected to vertices v0, v1 with hyperedge e0, to vertices v1, v3 with hyperedge e1, to vertices v3, v4 with hyperedge e2.\\
   Vertex v3 is connected to vertices v1, v2 with hyperedge e1, to vertices v2, v4 with hyperedge e2.\\
    Vertex v4 is connected to vertices v2, v3 with hyperedge e2.}
\end{myprompt}

\subsection{Examples of Questions for Different Tasks}
\label{app:prompt_task}
We provide representative question examples for each of the 12 tasks in \texttt{HyperGVL}, organized by difficulty levels. These examples are taken directly from the actual benchmark dataset.

\subsubsection{Level-1 Tasks}

\begin{myprompt}[label=prompt:vc]{Vertex Counting (VC)}
\textbf{Q:} How many vertices are in the hypergraph $\mathcal{G}$? List the answer after ``Ans:''.
\end{myprompt}

\begin{myprompt}[label=prompt:hec]{Hyperedge Counting (HEC)}
\textbf{Q:} How many hyperedges are in the hypergraph $\mathcal{G}$? List the answer after ``Ans:''.
\end{myprompt}

\begin{myprompt}[label=prompt:ne]{Neighbor (Ne)}
\textbf{Q:} What are the direct neighbors of vertex $v_4$ in hypergraph $\mathcal{G}$? (Neighbors = vertices sharing at least one hyperedge with $v_4$). List the answer after ``Ans:'' in the format \texttt{\{v1,v2,...\}} or ``No neighbors''.
\end{myprompt}

\subsubsection{Level-2 Tasks}

\begin{myprompt}[label=prompt:dvc]{Degree-specified Vertex Counting (DVC)}
\textbf{Q:} How many vertices have degree 3 in hypergraph $\mathcal{G}$? (Degree = number of hyperedges the vertex belongs to). List the answer after ``Ans:''.
\end{myprompt}

\begin{myprompt}[label=prompt:oec]{Order-specified HyperEdge Counting (OEC)}
\textbf{Q:} How many hyperedges have order 4 in hypergraph $\mathcal{G}$? (Order = number of vertices in the hyperedge). List the answer after ``Ans:''.
\end{myprompt}

\begin{myprompt}[label=prompt:ONe]{Order-filtered Neighbors (ONe)}
\textbf{Q:} What are the neighbors of vertex $v_5$ when only considering hyperedges with order $Eq 2$ in hypergraph $\mathcal{G}$? List the answer after ``Ans:'' in the format \texttt{\{v1,v2,...\}} or ``No n-neighbors''.
\end{myprompt}

\subsubsection{Level-3 Tasks}

\begin{myprompt}[label=prompt:osp]{Order-weighted Shortest Path (OSP)}
\textbf{Q:} What is the shortest path length from vertex $v_4$ to vertex $v_8$ in hypergraph $\mathcal{G}$, where each hyperedge's weight equals its order (number of vertices)? If no path exists, answer ``No path''. List the answer after ``Ans:''.
\end{myprompt}

\begin{myprompt}[label=prompt:omf]{Order-weighted Maximum Flow (OMF)}
\textbf{Q:} What is the estimated maximum flow from vertex $v_5$ to vertex $v_0$ in hypergraph $\mathcal{G}$, where each hyperedge's capacity equals its order? If no flow exists, answer ``0''. List the answer after ``Ans:''.
\end{myprompt}

\begin{myprompt}[label=prompt:ism]{Isomorphism Recognition (ISM)}
There are two hypergraphs: $H$ and $\mathcal{G}$.\newline
The description of $H$ is: \newline
The description of $\mathcal{G}$ is: \newline
\textbf{Q:} Are these two hypergraphs isomorphic? (Two hypergraphs are isomorphic if there exists a vertex relabeling that transforms one into the other). List the answer after ``Ans:'' in the format [Yes/No].
\end{myprompt}

\subsubsection{Level-4 Tasks}

\begin{myprompt}[label=prompt:3cl]{Hypergraph 3-Coloring (3-CL)}
\textbf{Q:} Please provide a 3-coloring strategy such that each hyperedge contains nodes with at least 2 different colors (assign each vertex a color from \{c0, c1, c2\}). List the answer after ``Ans:'' as ``Coloring:[v0:c0,v1:c1,...]''.
\end{myprompt}

\begin{myprompt}[label=prompt:shc]{Strict Hypercycle (SHC)}
\textbf{Q:} Please identify a strict hypercycle in the hypergraph $\mathcal{G}$ (A strict hypercycle is a sequence of hyperedges $e_1,e_2,...,e_k$ where adjacent hyperedges share exactly one vertex, i.e., $|e_i \cap e_{i+1}| = 1$, and $|e_k \cap e_1| = 1$, forming a closed loop). List the hypercycle after ``Ans:'' as ``Cycle:[e0,e1,...]''.
\end{myprompt}

\begin{myprompt}[label=prompt:hhm]{Hypergraph Hamiltonian Path (HHM)}
\textbf{Q:} Please provide a valid Hamiltonian path from v1 to v0.
(Hamiltonian path = path visiting all vertices exactly once). List the answer after ``Ans:'' as ``Path:[e0,e1,...]''.
\end{myprompt}

Despite there is format guidance in the prompt, we perform manual review to accept factually correct answers with minor formatting deviations, ensuring evaluation fairness.

\section{HyperGVL Evaluation Details}
\label{app:evaluation}
For all evaluated LVLMs and LLMs, we set the temperature parameter to 0.8 to introduce a controlled level of randomness during generation, which allows us to assess both the stability and consistency of model outputs across different runs. We adopt nucleus sampling with a top-$p$ value of 0.95 and a maximum generation length sufficient to cover all task-specific output requirements. All experiments were conducted on a compute cluster equipped with 64 NVIDIA RTX 5090 GPUs. Model inference was executed using mixed-precision computation to balance efficiency and numerical stability. The evaluation pipeline was parallelized across GPUs to ensure consistent runtime conditions for all models.

\section{Hypergraph Node Classification Details}
\label{app:real}

\subsection{Dataset Statistics}
In Sec. \ref{sec:router}, we demonstrate the out-of-domain generalizability of \texttt{WiseHyGR} on real-world hypergraph node classification under the zero-shot setting. The tested hypergraphs come from multiple sources, including co-authorship hypergraphs and co-citation hypergraphs. The dataset statistics are provided in Tab.\ref{tab:realset}.

\begin{itemize}
    \item Co-authorship hypergraphs: Cora-CA \footnote{\url{https://people.cs.umass.edu/ mccallum/data.html}} and DBLP-CA\footnote{\url{https://aminer.org/lab-datasets/citation/DBLP-citation-Jan8.tar.bz}} \cite{yadati2019hypergcn}, where vertices represent articles and hyperedges denote the co-author relationship of articles that share a same author.
    \item Co-citation hypergraphs: Cora, Citeseer, and Pubmed\footnote{\url{https://linqs.soe.ucsc.edu/data}} \cite{yang2016revisiting}, where vertices are articles and hyperedges connect all articles that are cited by the same article.

\end{itemize}

\subsection{Experimental Setups}
From each dataset, we sample 1K hypergraph nodes for classification. To ensure these nodes (i.e., vertices) form a connected hypergraph, we initiate a random walk from a target node, collecting up to $n$ nodes. We set $n=40$ in our experiments to provide sufficient structural information to the LVLMs while minimizing noise and information congestion from excessively long-range dependencies and overly broad scopes. The performances shown in Fig.\ref{fig:router_out} are averaged over three trials to ensure robust comparisons. Zero-shot hypergraph node classification is conducted using prompts exemplified below, with categorization descriptions generated by GPT-4o \cite{gpt4o}. The node attributes are the manually retrieved article abstracts.

\begin{myprompt}[label=prompt:hypergraph-class]{Hypergraph Node Classification Prompts}
    \textbf{Prompt: }\textit{Classify node v0 in the hypergraph.\\
    Hypergraph Structure: \{Textual Hypergraph Representation\}\\
    Known labels in the subgraph:
    \begin{itemize}
        \item v1: Probabilistic Methods
        \item v2: Reinforcement Learning 
        \item ...
    \end{itemize}
    Target Node Content: \{Paper Abstract\}\\
    Categories:
    \begin{itemize}
        \item Genetic Algorithms: ...
        \item Probabilistic Methods: ...
        \item ...
    \end{itemize}
    Based on the hypergraph structure, known labels of other nodes, and the target node's abstract content, which category does node v0 most likely belong to?\\
    Answer with the category name only.}
\end{myprompt}

\end{document}